\pdfoutput=1
\documentclass{article}

\usepackage[preprint]{neurips_2021}

\usepackage[utf8]{inputenc} 
\usepackage[T1]{fontenc}    
\usepackage{hyperref}       
\usepackage{url}            
\usepackage{booktabs}       
\usepackage{amsfonts}       
\usepackage{nicefrac}       
\usepackage{microtype}      
\usepackage{xcolor}         

\usepackage{tabularx}
\usepackage{thm-restate}
\usepackage[ruled,vlined]{algorithm2e}
\SetAlFnt{\footnotesize}
\SetAlCapFnt{\footnotesize}
\SetAlCapNameFnt{\normalsize}
\usepackage{amsmath,amsthm,amssymb}
\usepackage{bbm, bm}
\usepackage{mathtools}
\usepackage{float}
\usepackage{caption}

\ifx\theorem\undefined{\newtheorem{theorem}{Theorem}}\fi
\ifx\proposition\undefined{\newtheorem{proposition}{Proposition}}\fi
\ifx\corollary\undefined{\newtheorem{corollary}{Corollary}}\fi
\ifx\lemma\undefined{\newtheorem{lemma}{Lemma}}\fi

\ifx\definition\undefined{\newtheorem{definition}{Definition}}\fi
\ifx\example\undefined{\newtheorem{example}{Example}}\fi
\ifx\assumption\undefined{\newtheorem{assumption}{Assumption}}\fi
\ifx\remark\undefined{\newtheorem{remark}{Remark}}\fi

\DeclareMathOperator{\E}{\mathbb{E}}
\DeclareMathOperator{\KL}{\mathbb{KL}}
\DeclareMathOperator{\parallelbars}{%
  \,\|\,%
}
\renewcommand{\eqref}[1]{(\ref{#1})}

\DeclareMathOperator{\Bern}{\operatorname{Bern}}

\DeclareMathOperator{\x}{\mathbf{x}}

\DeclareMathOperator{\bfa}{\mathbf{a}}
\DeclareMathOperator{\bfd}{\mathbf{d}}

\DeclareMathOperator{\bfalpha}{\bm{\alpha}}
\DeclareMathOperator{\bfgamma}{\bm{\gamma}}
\DeclareMathOperator{\bfbeta}{\bm{\beta}}

\DeclareMathOperator{\z}{\mathbf{z}}
\DeclareMathOperator{\s}{\mathbf{s}}

\usepackage{subcaption}
\usepackage[export]{adjustbox}

\title{Unsupervised Causal Binary Concepts Discovery with VAE for Black-box Model Explanation}

\author{%
  Thien Q. Tran \\
  University of Tsukuba, Riken AIP \\
  \texttt{thientquang@mdl.cs.tsukuba.ac.jp} \\
  \And
  Kazuto Fukuchi\\
  University of Tsukuba, Riken AIP \\
  \texttt{fukuchi@mdl.cs.tsukuba.ac.jp} \\
  \And
  Youhei Akimoto \\
  University of Tsukuba, Riken AIP \\
  \texttt{akimoto@mdl.cs.tsukuba.ac.jp} \\
  \And
  Jun Sakuma \\
  University of Tsukuba, Riken AIP \\
  \texttt{jun@mdl.cs.tsukuba.ac.jp} \\
}

\begin{document}
\maketitle

\begin{abstract}
  We aim to explain a black-box classifier with the form: `data X is classified as class Y because X \textit{has} A, B and \textit{does not have} C' in which A, B, and C are high-level concepts. The challenge is that we have to discover in an unsupervised manner a set of concepts, i.e., A, B and C, that is useful for the explaining the classifier. We first introduce a structural generative model that is suitable to express and discover such concepts. We then propose a learning process that simultaneously learns the data distribution and encourages certain concepts to have a large causal influence on the classifier output. Our method also allows easy integration of user's prior knowledge to induce high interpretability of concepts. Using multiple datasets, we demonstrate that our method can discover useful binary concepts for explanation.
\end{abstract}

\section{Introduction\label{ch:intro}}
\begin{figure}[h]
  \vspace{-1.5em}
  \captionsetup{aboveskip=5pt}
  \captionsetup[subfigure]{aboveskip=2pt}
  \centering
  \begin{subfigure}[b]{0.45\linewidth}
    \centering
    \includegraphics[width=0.85\linewidth]{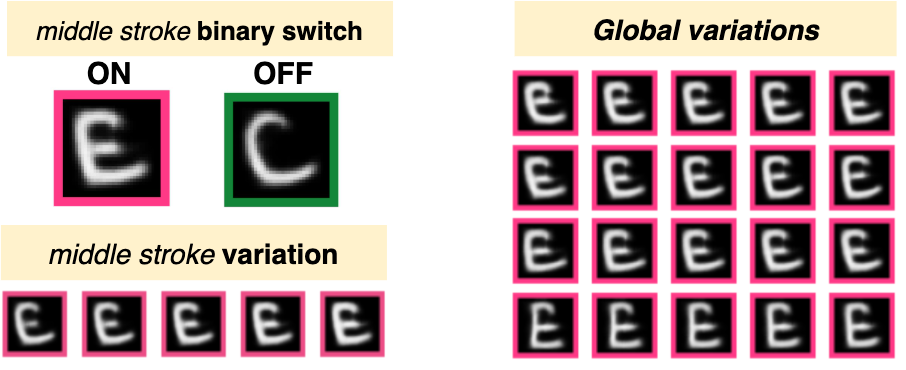}
    \caption{Illustration of proposed concepts}\label{fig:illustration}
  \end{subfigure}%
  \begin{subfigure}[b]{0.22\linewidth}
    \centering
    \includegraphics[width=0.9\linewidth]{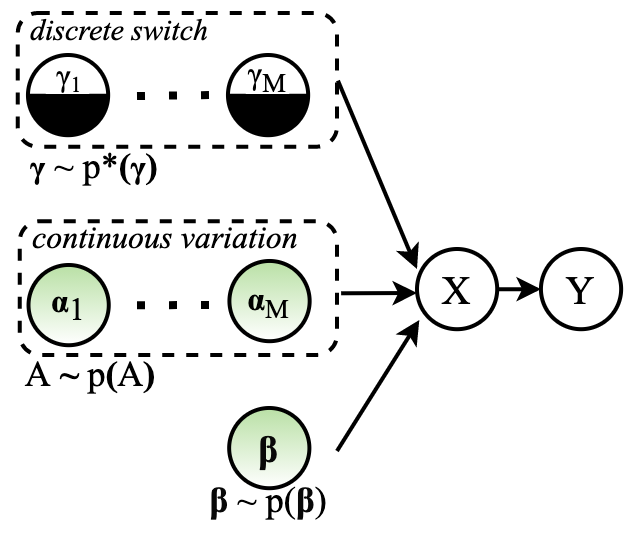}
    \caption{Causal DAG}\label{fig:proposed_dag}
  \end{subfigure}%
  \begin{subfigure}[b]{0.35\linewidth}
    \centering
    \includegraphics[width=0.9\linewidth]{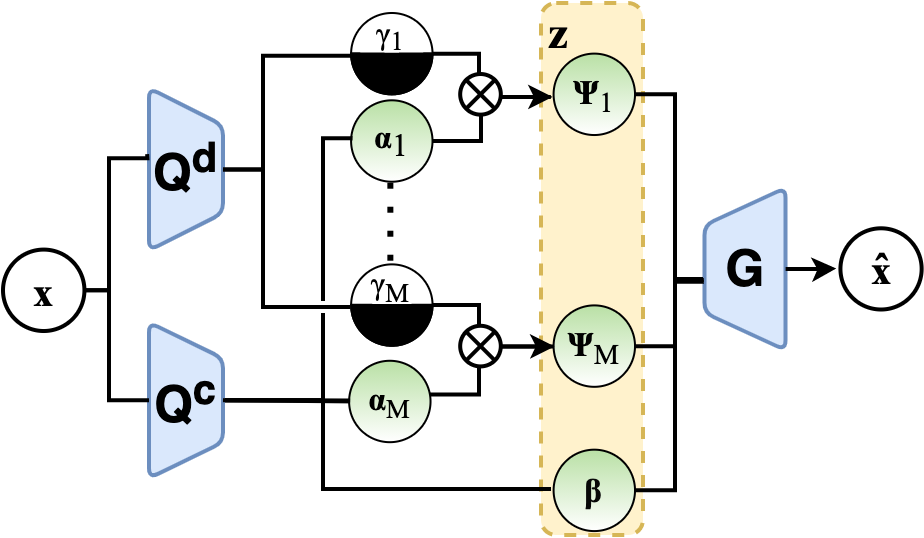}
    \caption{VAE model}\label{fig:proposed_vae}
  \end{subfigure}%
  \caption{(a) Binary concept \textit{middle stroke} and some global variants. Border color indicates the classifier output. (b, c) The proposed VAE model and the causal DAG}
  \vspace{-0.5em}
\end{figure}

Deep neural network has been recognized as the state-of-the-art model for various tasks. As they are being applied in more practical applications, there is an arising consensus that these models need to be explainable, especially in high-stake domains. Various methods are proposed to solve this problem, including building interpretable model and post-hoc methods that explain trained black-box models. We focus on the post-hoc approach and propose a novel causal concept-based explanation framework.

We are interested in an explanation that uses the symbolic expression: `data X is classified as class Y because X \textit{has} A, B and \textit{does not have} C' where A, B, and C are high-level concepts. From the linguistic perspective, our explanation communicates using \textit{nouns} and their \textit{part-whole relation}, i.e., the semantic relation between a part and the whole object. In many classification tasks, especially image classification, the predictions relied on binary components; for example, we can distinguish a panda from a bear by its white patched eyes or a zebra from a horse by its stripe. This is also a common way humans use to classify categories and organize knowledge \cite{Gardenfors2014-rq}. Thus, an explanation in this form should excel in providing human-friendly and organized insights into the classifier, especially for tasks that involve higher-level concepts such as checking the alignment of the black-box model with experts. From now on, we refer to such a concept as \textit{binary concept}.

Our method employs three different notions in the explanation: \textit{causal binary switches}, \textit{concept-specific variants} and \textit{global variants}. We illustrate these notions in Figure \ref{fig:illustration}. First, \textit{causal binary switches} and \textit{concept-specific variants}, that come in pair, represent different binary concepts. In particular, \textit{causal binary switches} control the presence of each binary concept in a sample. Alternating this switch, i.e., removing or adding a binary concept to a sample, affects the prediction of that sample (e.g., removing the middle stroke turns E to C). In contrast, \textit{concept specific variants}, whose each is tied to a specific binary concept, express different variants within a binary concept that do not affect the prediction (e.g., changing the length of the middle stroke does not affect the prediction). Finally, \textit{global variants}, which are not tied to specific binary concepts, represent other variants that do not affect the prediction (e.g., skewness).

\newdimen\figrasterwd
\figrasterwd\textwidth
\begin{figure}[t]
  \captionsetup{aboveskip=5pt}
  \captionsetup[subfigure]{aboveskip=2pt}
  \centering
  \parbox{\figrasterwd}{
    \parbox[b]{.23\figrasterwd}{%
      \centering
      \subcaptionbox{Saliency methods\label{fig:saliency-maps}}{\includegraphics[width=0.9\hsize]{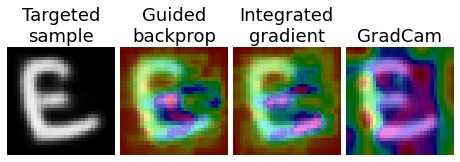}}
      \subcaptionbox{VSC model\label{fig:mixed-vae}}{\includegraphics[width=0.9\hsize]{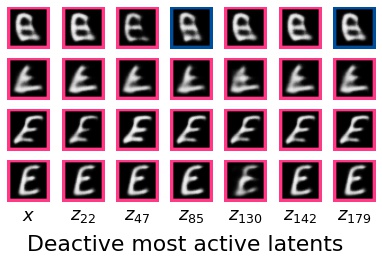}}
    }
    \parbox[b]{.23\figrasterwd}{%
      \subcaptionbox{O’Shaughnessy et al.\label{fig:cc-vae}}{
        \begin{subfigure}[b]{1\linewidth}
          \centering
          \includegraphics[width=0.9\linewidth]{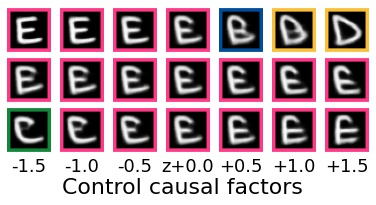}
          \includegraphics[width=0.9\linewidth]{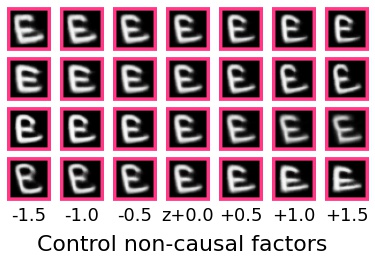}
        \end{subfigure}
      }
    }
    \parbox[b]{.29\figrasterwd}{%
      \centering
      \subcaptionbox{Proposed (causal factors)\label{fig:proposed-discrete}}{\includegraphics[width=1.0\hsize]{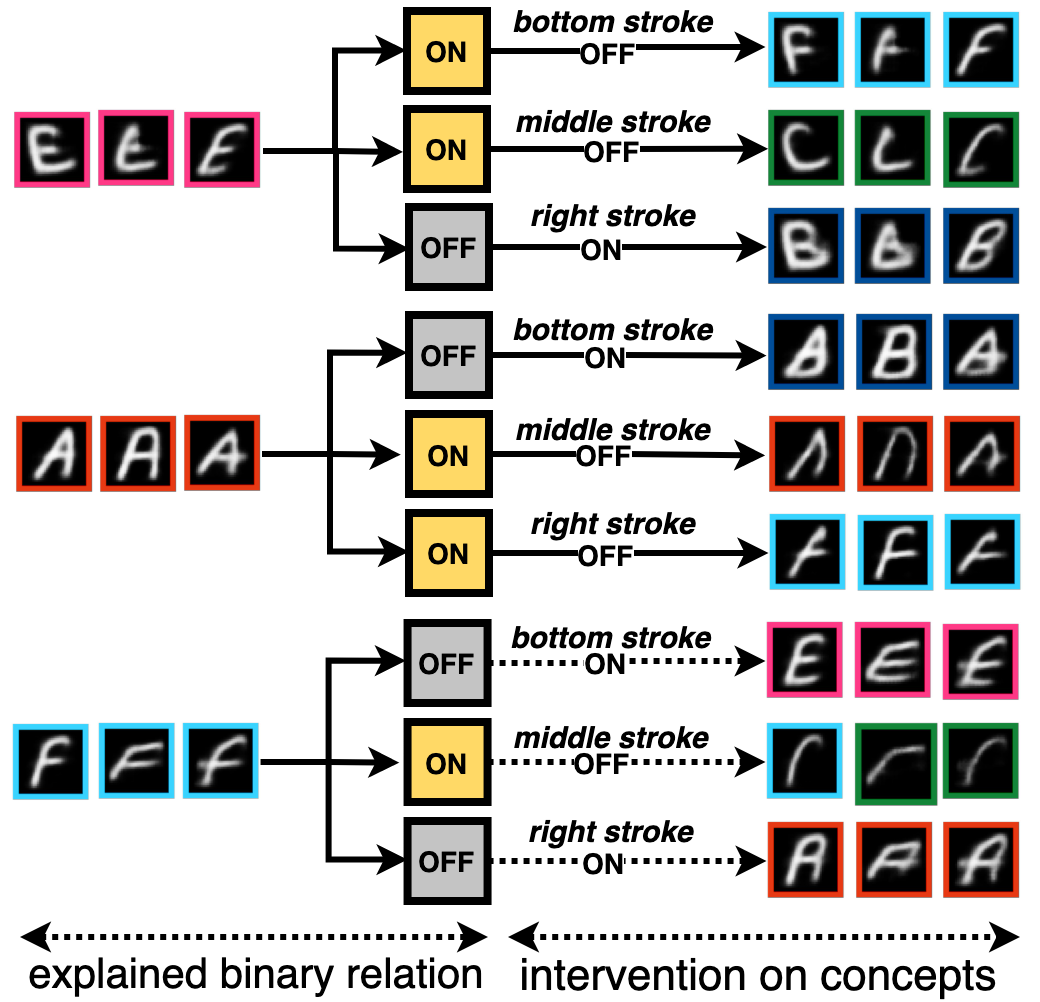}
      }
    }
    \parbox[b]{.23\figrasterwd}{%
      \subcaptionbox{Proposed (non-causal)\label{fig:proposed-continuous}}{
        \begin{subfigure}[b]{1\linewidth}
          \centering
          \includegraphics[width=0.9\linewidth]{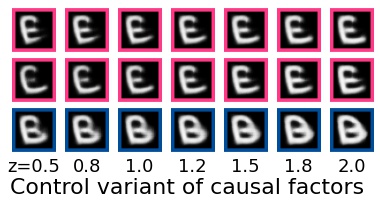}
          \includegraphics[width=0.9\linewidth]{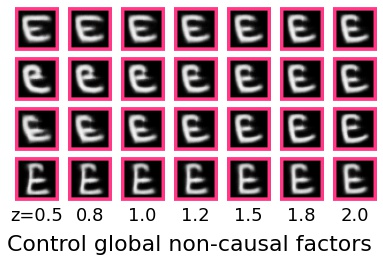}
        \end{subfigure}
      }
    }
    \caption{Explanation methods for a letter classifier. (a) Saliency-based methods. (b) Disabling the most active latents of class E in VSC model. (c) Controlling the causal and non-causal factors in O’Shaughnessy et al. (d, e) Proposed method: (d) Encoded binary relation of discovered concepts and their intervention results; (e) variants within each concept and other variants of the whole letter.\label{fig:demonstration}}}
  \vspace{-1.5em}
\end{figure}

Our goal is to discover a set of binary concepts that can explain the classifier using their binary switches in an unsupervised manner. Similar to some existing works, to construct conceptual explanations, we learn a generative model that maps each input into a low-dimensional representation in which each factor encodes an aspect of the data. There are three main challenges in achieving our goal. (1) It requires an adequate generative model to express the binary concepts, including the binary switches and the variants within each concept. (2) The discovered binary concepts must have a large causal influence on the classifier output. That is, we avoid finding confounding concepts, which correlate with but do not cause the prediction. For example, the \textit{sky} concept appears frequently in \textit{plane}'s images but may not cause the prediction of \textit{plane}. (3) The explanation must be interpretable and provide useful insights. For example, a concept that entirely replaces a letter E with a letter A has a large causal effect. However, such a concept does not provide valuable knowledge due to lack of interpretability.

In Figure \ref{fig:proposed-discrete} and \ref{fig:proposed-continuous}, we demonstrate an explanation discovered by the proposed method for an classifier for six letters: A,B,C,D,E and F. Our method successfully discovered the concepts of \textit{bottom stroke}, \textit{middle stroke} and \textit{right stroke} which effectively explains the classifier. In Figure \ref{fig:proposed-discrete}, we show the encoded binary switches and their interventions result. From the top figure, we can explain that: this letter is classified as E because it \textit{has} a \textit{bottom stroke} (otherwise it is F), a \textit{middle stroke} (otherwise it is C), and it \textit{does not have} a \textit{right stroke} (otherwise it is B). We were also able to distinguish the variant within each concept in (Figure \ref{fig:proposed-continuous} top) with the global variant (Figure \ref{fig:proposed-continuous} bottom). A full result with explanation for other letters is shown in Section \ref{ch:experiment}.

To the best of our knowledge, no existing method can discover binary concepts that fulfill all of these requirements. Saliency methods such as Guided Backprop \cite{Springenberg2014-dc}, Integrated Gradient \cite{Sundararajan2017-gt} or GradCam \cite{Selvaraju2020-ap} only show feature importance but do not explain why (Figure \ref{fig:saliency-maps}). Some generative models which use binary-continuous mixed latents for sparse coding, such as VSC \cite{Tonolini2020-cm}, IBP-VAE \cite{Gyawali2019-wm}, PatchVAE \cite{Gupta2020-zc}, can support binary concepts. However, they do not necessarily discover binary concepts that are useful for explanation, in both causality and interpretability (Figure \ref{fig:mixed-vae}). Recently, \citet{OShaughnessy2020-yz} proposed a learning framework that encourages the causal effect of certain latent factors on the classifier output to learn a latent representation that has causality on the prediction. However, their model can not disentangle binary concepts and can be hard to interpret, especially for multiple-class tasks. For example, a single concept changes the letter E to multiple other letters (Figure \ref{fig:cc-vae}), which would not give any interpretation on how this latent variable affects prediction.

Our work has the following contributions: (1) We introduce the problem of discovering binary concepts for the explanation. Then, we propose a structural generative model for constructing binary concept explanation, which can capture the binary switches, concept-specific variants, and global variants. (2) We propose a learning process to simultaneously learn the data distribution while encouraging the causal influence of the binary switches. Although typically VAE models encourage the independence of factors for meaningful disentanglement, such an assumption is inadequate for discovering useful causal concepts which are often mutually correlated. Our learning process, which considers the dependence between binary concepts, can discover concepts with more significant causality. (3) To avoid the concepts that have causality but no interpretability, the proposed method allows an easy way to implement user's preference and prior knowledge as a regularizer to induce high interpretability of concepts. (4) Finally, we demonstrate that our method succeeds in discovering interpretable binary concepts with causality that are useful for explanation with multiple datasets.

\section{Related Work\label{ch:pastwork}}
Our method can be categorized as a concept-based method that explains using high-level aspects of data. The definition of \textit{concept} are various, e.g., a direction in the activation space \cite{Kim2018-wr, Ghorbani2019-un}, a prototypical activation vector \cite{Yeh2019-lr} or a latent factor of a generative model \cite{OShaughnessy2020-yz, Goyal2020-se}. We remark that this notion of concept should depend on the data and the explanation goal. Some works defined the concepts beforehand using additional data. When this side-information is not given, one needs to discover useful concepts for the explanation, e.g., \citet{Ghorbani2019-un} used segmentation and clustering, \citet{Yeh2019-lr} retrained the classifier with a prototypical concept layer, \citet{OShaughnessy2020-yz} learned the generative model with a causal objective.

A generative model such as VAE can provide a concept-based explanation as it learns a latent presentation $\z$ that captures different aspects of the data. However, \citet{Bauer_undated-ht} shows that disentangled representations in a fully unsupervised manner are fundamentally impossible without inductive bias. A popular approach is to augment the VAE loss with a regularizer \cite{noauthor_undated-vd, Burgess2018-ce}. Another approach is to incorporate structure into the representation\cite{Choi2020-av, Ross2021-rf, Tonolini2020-cm, Gupta2020-zc}. Although these methods can encourage disentangled and sparse representation, the learned representations are not necessarily interpretable and have causality  on the classifier output.

We pursue an explanation that has causality. A causal explanation is helpful as it can avoid attributions and concepts that only correlate with but do not causes the prediction. Previous works have attempted to focus on causality in various ways. For example, \citet{Schwab2019-cm} employed Granger causality to quantify the causal effect of input features, \citet{Parafita2019-lx} evaluated the causality of latent attributions with a prior known causal structure, \citet{Narendra2018-kc} evaluated the causal effect of network layers, and \citet{Kim2019-kp} learned an interpretable model with a causal guarantee. To the best of our knowledge, no existing works can explain using binary concepts that fulfill the three requirements we discussed.

\section{Preliminaries\label{ch:background}}
\subsection{Variational Autoencoder\label{ch:background-vae}}
Our explanation is build upon the VAE framework proposed by \citet{Kingma2014-bb}. VAE model assumes a generative process of data in which a latent $\z$ is first sampled from a prior distribution $p(\z)$, then the data is generated via a conditional distribution $p(\x \mid \z)$. Typically, due to the intractability, a variational approximation $q(\z \mid \x)$ of the intractable posterior is introduced and the model is then learned using the evidence lower bound (ELBO) as
\begin{align}
  \begin{split}
    \mathcal{L}_\text{VAE}(\x) = &-\E_{\z \sim q(\z \mid \x)}[\log p(\x \mid \z)] + \KL[q(\z \mid \x) \parallelbars p(\z)].
  \end{split}
\end{align}
Here, $q(\z \mid \x)$ is the encoder that maps the data to the latent space and $p(\x \mid \z)$ is the decoder that maps the latents to the data space. Commonly, $q(\z \mid \x)$ and $p(\x \mid \z)$ are parameterized as neutral networks $Q(\z \mid \x)$ and $G(\x \mid \z)$, respectively. The common choice for $q(\z \mid \x)$ is a factorized Gaussian encoder $q(\z \mid \x) = \prod_{p=1}^P \mathcal{N}(\mu_i, \sigma_i^2)$ where $(\mu_1, \dots, \mu_P, \sigma_1, \dots, \sigma_P,) = Q(\x)$. The common choice for the $p(\z)$ is a multi-variate normal distribution $\mathcal{N}(0, \mathcal{I})$ with zero mean and identity covariant. Then, the first term can be trained using L2 reconstruction loss, while the KL-divergence terms are trained using the reparameterization trick.

\subsection{Information Flow\label{ch:background-if}}
Next, we introduce the measure we use to quantify the causal influence of the learned representation on the classifier output. We adopt Information Flow, which defines the causal strength using Pearl's do calculus \cite{Pearl2009-pa}. Given a causal directional acyclic graph $G$, Information Flow quantify the statistical influence using the conditional mutual information on the interventional distribution:
\begin{definition}[Information flow from $U$ to $V$ in a directed acyclic graph $G$\cite{Ay2008-km}]\label{def:iflow}
  Let $U$ and $V$ be disjoint subsets of nodes. The information flow $I(U \rightarrow V)$ from $U$ to $V$ is defined by
  \begin{align}
    I(U \rightarrow V) = \int_U p(u) \int_V p(v | \text{do}(u)) \log \frac{p(v | \text{do}(u))}{\int_{u'} p(u')p(v | \text{do}(u'))du'} dV dU,
  \end{align}
  where $\text{do}(u)$ represents an intervention that fixes u to a value regardless of the values of its parents.
\end{definition}
\citet{OShaughnessy2020-yz} argued that compared to other metrics such as average causal effect (ACE) \cite{Holland1988-xs}, analysis of variance (ANOVA) \cite{Lewontin1974-wq}, information flow is more suitable to capture complex and nonlinear causal dependence between variables.

\section{Proposed method\label{ch:method}}
We aim to discover a set of binary concepts $\mathcal{M} = \{m_0, m_1, \dots, m_M\}$ with causality and interpretability that can explain the black-box classifier $f: \mathcal{X} \rightarrow \mathcal{Y}$. Inspired by \citet{OShaughnessy2020-yz}, we employs a generative model to learn the data distribution while encouraging the causal influence of certain latent factors. In particular, we assume a causal graph in Figure \ref{fig:proposed_dag}, in which each sample $\x$ is generated from a set of latent variables, including $M$ pairs of \textit{a binary concept} and \textit{a concept-specific variant} $\{\gamma_i, \bfalpha_i\}_{i=1}^M$, and a \textit{global variants} $\bfbeta$. As we want to explain the classifier output (i.e., node $y$ in Figure \ref{fig:proposed_dag}) using the \textit{binary switches} $\{\gamma_i\}$, we expect that $\{\gamma_i\}$ has a large causal influence on $y$.

Our proposed learning objective consists of three components, which corresponds to our three requirements: a VAE objective $\mathcal{L}_\text{VAE}$ for learning the data distribution $p(\x)$, a causal effect objective $\mathcal{L}_\text{CE}(X)$ for encouraging the causal influence of $\{\gamma_i\}$ on classifier output $y$, and an user-implementable regularizer $\mathcal{L}_\text{R}(\x)$ for improving the interpretability and consistency of discovered concepts:
\begin{align}
  \mathcal{L}(X) = \frac{1}{|X|}\sum_{\x \in X} \left[\mathcal{L}_\text{VAE}(\x) + \lambda_\text{R}\mathcal{L}_\text{R}(\x)\right] + \lambda_\text{CE} \mathcal{L}_\text{CE}(X). \label{eq:all}
\end{align}

\subsection{VAE model with binary concepts\label{ch:vae}}
To represent the binary concepts, we employ a structure in which each binary concept $m_i$ is presented by a latent variable $\bm{\psi}_i$, which is further controlled by two factors: a binary concept switch latent variable $\gamma_i$ (concept switch for short) and a continuous latent variable representing concept-specific variants $\bfalpha_i$ (concept-specific variant for short) as $\bm{\psi}_i = \gamma_i \cdot \bfalpha_i$ where $\gamma_i = 1$ if concept $m_i$ is on and $gamma_i=0$ otherwise. Here, the \textit{concept switch} $\gamma_i$ controls if the concept $m_i$ is activated in a sample, e.g., if the bottom stroke is appeared in a image (Figure \ref{fig:proposed-discrete}). On the other hand, the \textit{concept-specific variant} $\bfalpha_i$ controls the variant within the concept $m_i$, e.g., the length of the bottom stroke (Figure \ref{fig:proposed-continuous}, top). In addition to the \textit{concept-specific variants} $\{\bfalpha_i\}$ whose effect is limited to a specific binary concept, we also allow a \textit{global variant} latent $\bfbeta$ to capture other variants that do not necessarily have causality, e.g., skewness (Figure \ref{fig:proposed-continuous}, bottom). Here, disentangling concept-specific and global variants is important for assisting user in understanding discovered binary concepts.

The way we represent binary concepts is closely related to the spike-and-slab distribution, which is used in Bayesian variable selection \cite{George_undated-ah} and sparse coding \cite{Tonolini2020-cm}. Unlike these models, whose number of discrete-continuous factors is often large, our model uses only a small number of binary concepts with a multi-dimensional global variants $\beta$. Our intuition is that in many cases, the classification can be made by combining a small number of binary concepts.

\textbf{Input encoding.} 
For the discrete components, we use a network $Q^d(\x)$ to parameterize $q(\bfgamma \mid \x)$ as $q(\bfgamma \mid \x) = \prod_{i=1}^M q(\gamma_i \mid \x) = \prod_{i=1}^M \text{Bern}(\gamma_i; \pi_i)$  where $(\pi_1, \dots, \pi_M) = Q^d(\x)$. For the continuous components, letting $A = (\bfalpha_1 , \bfalpha_2, \dots, \bfalpha_M)$, we use a network $Q^c(\x)$ to parameterize $q(A, \bfbeta \mid \x)$ as $q(A, \bfbeta \mid \x) = \left[\prod_{i=1}^M q\left(\bfalpha_i \mid \x\right)\right] q(\bfbeta \mid \x)$. Here, $q(\bfalpha_i \mid \x) = \mathcal{N}^\text{fold}_\delta(\bfalpha_i; \mu_i, \text{diag}(\sigma_i))$, $q(\bfbeta \mid \x) = \mathcal{N}^\text{fold}_\delta(\bfbeta; \mu_{\bfbeta}, \text{diag}(\sigma_{\bfbeta}))$ and $(\mu_1, \dots, \mu_M, \mu_{\bfbeta}, \sigma_1, \dots, \sigma_M, \sigma_{\bfbeta}) = Q^c(\x)$. Here, we employ the $\delta\text{-Shifted}$ Folded Normal Distribution $\mathcal{N}^\text{fold}_\delta(\mu, \sigma^2)$ for continuous latents, which is the distribution of $|x| + \delta$ with a constant hyper-parameter $\delta > 0$ where $x \sim \mathcal{N}(\mu, \sigma ^2)$. In all of our experiments, we adopted $\delta=0.5$. We choose not the standard Normal Distribution but the $\delta\text{-Shifted}$ Folded Normal Distribution because it is more appropriate for the causal effect we want to achieve. The implementation of $\mathcal{N}^\text{fold}_\delta(\mu, \sigma^2)$ can simply be done by adding the absolute and shift operation to the conventional implementation of $\mathcal{N}(\mu, \sigma ^2)$. We discuss in detail this design choice in Appendix \ref{appendix:fold-shilf}.

\textbf{Output decoding.} Next, given $q(\bfgamma \mid \x)$ and $q(A, \bfbeta \mid \x)$, we first sample the concept switches $\{\hat{d}_i\}$, the concept variants $\{\hat{\bfalpha}_i\}$ and the global variants $\bfbeta$ from their posterior, respectively.
Using these sampled latents, we construct an aggregated representation $\hat{\z} = (\bm{\psi}_1, \dots, \bm{\psi}_M, \hat{\bfbeta})$ using the binary concept mechanism in which $\bm{\psi}_i$ is the corresponding part for concept $m_i$, i.e., $\bm{\psi}_i = \gamma_i \times \bfalpha_i$. If concept $m_i$ is on, we let $\hat{d}_i=1$ so that $\bm{\psi}_i$ can reflect the concept-specific variant $\hat{\bfalpha}_i$. Otherwise, when the concept $m_i$ is off, we assign $\hat{d}_i=0$. We refer to $\hat{\z}$ as the \textit{conceptual latent code}. Finally, a decoder network takes $\hat{\z}$ and generate the reconstruction $\hat{x}$ as $\hat{x} \sim G(\x \mid \hat{\z}) \text{ where } \hat{\z} = (\bm{\psi}_1, \dots, \bm{\psi}_M, \hat{\bfbeta}).$

\textbf{Learning process.} We use the maximization of evidence lower bound (ELBO) to jointly train the encoder and decoder. We assume the prior distribution for continuous latents to be $\delta\text{-shifted}$ Folded Normal distribution $\mathcal{N}^\text{fold}_\delta(0, \mathcal{I})$ with zero-mean and identity covariance. Moreover, we assume the prior distribution for binary latents to be a Bernoulli distribution $\text{Bern}(\pi_{\text{prior}})$ with prior $\pi_{\text{prior}}$. The ELBO for our learning process can be written as:
\begin{align}
  \begin{split}
    \mathcal{L}_\text{VAE}(\x) &= -\E_{\hat{\z} \sim Q^{\{c,d\}}(\z \mid \x)} \left[ \log G\left(\x \mid \hat{\z} \right)\right] + \lambda_2 \left[ \frac{1}{M} \sum_{i=1}^{M} \KL\left(q \left(\gamma_i \mid \mathbf{x}\right) \parallelbars \Bern\left(\pi_i \right)\right) \right] \\
    &+ \lambda_1 \left[ \KL \left(q\left(\bfbeta \mid \mathbf{x}\right) \parallelbars \mathcal{N}^\text{fold}_\delta\left(0,\mathcal{I}\;\right)\right) + \frac{1}{M} \sum_{i=1}^{M} \KL \left(q\left(\bfalpha_i \mid \mathbf{x}\right) \parallelbars \mathcal{N}^\text{fold}_\delta\left(0,\mathcal{I}\;\right)\right) \right]. \label{eq:binary_vae}
  \end{split}
\end{align}
For the Bernoulli distribution, we use its continuous approximation, i.e., the relaxed-Bernoulli \cite{Maddison2016-ys} in the training process.

\subsection{Encouraging causal effect of binary switches\label{ch:causal}}
We expect the binary switches $\bfgamma$ to have a large causal influence so that they can effectively explain the classifier. To measure the causal effect of $\bfgamma$ on the classifier output $Y$, we employ the causal DAG in Figure \ref{fig:proposed_dag} and adopt \textit{information flow} (Definition \ref{def:iflow}) as the causal measurement. Our DAG employs an assumption that is fundamentally different from those of standard VAE models. Specifically, the standard VAE model and also \citet{OShaughnessy2020-yz} assumes the independence of latent factors, which is believed to encourage meaningful disentanglement via a factorized prior distribution. We claim that because \textit{useful concepts for explanation} often causally depend on the class information and thus are not independent of each other, such an assumption might be inadequate for discovering valuable causal concepts. For example, in the letter E, the middle and the bottom strokes are causally related to the recognition of the letter E, and corresponding binary concepts are mutually correlated. Thus, employing the VAE's factorized prior distribution in estimating information flow might lead to a large estimation error and prevent discovering valuable causal concepts.

Instead, we employ a prior distribution $p^*(\bfgamma)$ that allows the correlation between causal binary concepts. Our method iteratively learns the VAE model and use the current VAE model to estimates the prior distribution $p^*(\bfgamma)$ which most likely generates the user's dataset. This empirical estimation of $p^*(\bfgamma)$ is then used to evaluate the causal objective in Eq. \eqref{eq:all}. Assuming $X$ is a set of i.i.d samples from data distribution $p(\x)$, we estimate $p^*(\bfgamma)$ as
\begin{align}
  p^*(\bfgamma) \approx \int_{\x} p^*(\bfgamma \mid \x)p(\x) d\x \approx \frac{1}{|X|}\sum_{\x \in X} p(\bfgamma \mid \x) \approx \frac{1}{|X|}\sum_{\x \in X} \prod_{i=1}^M q(\gamma_i \mid \x) \label{eq:estimate-prior}
\end{align}
In the last line, $p(\bfgamma \mid \x)$ is replaced with the variational posterior $q(\bfgamma \mid \x)$ of VAE model. Here, the factorized posterior $q(\bfgamma \mid \x)$ only assumes the independence between latents conditioned on a sample but does not imply the independence of binary switches in $p^*(\bfgamma)$. We note that we do not aim to learn the dependence between concepts but only expect that $p^*(\bfgamma)$ properly reflects the dependence between binary concepts that appears in the dataset $X$ for a better evaluation of causal effect. We experimentally show in Subsection \ref{ch:experiment-quantitative} that using the estimation of $p^*(\bfgamma)$ results in a better estimation for the causal effect on dataset $X$ and more valuable concepts for the explanation. We showed that in the proposed DAG, information flow $I(\bfgamma \rightarrow Y)$ coincides with mutual information $I(\bfgamma; Y)$.
\begin{proposition}[Coincident of Information Flow and Mutual Information in proposed DAG]\label{pro:mi-as-iflow}
  The information flow from $\bfgamma$ to Y in the DAG of Figure \ref{fig:proposed_dag} coincides with the mutual information between $\bfgamma$ and $Y$. That is, $I(\bfgamma \rightarrow Y) = I(\bfgamma; Y) = \E_{\bfgamma, Y} \left[\frac{p^*(\bfgamma)p(Y \mid \bfgamma)}{p^*(\bfgamma)p(Y)}\right]$.
\end{proposition}

We prove Proposition \ref{pro:mi-as-iflow} in Appendix \ref{appendix:proof}. The detailed algorithm for estimating $I(\bfgamma; Y)$ is described in Appendix \ref{appendix:algorithm}. As we want to maximize $I(\bfgamma; Y)$, we rewrite it as a loss term $\mathcal{L}_\text{CE} = -I(\bfgamma; Y)$ and optimize it together with the learning of VAE model.

\subsection{Integrating user preference for concepts\label{ch:preference}}
Finally, we discuss the integration of user's preferences or prior knowledge for inducing high interpretability of concepts. A problem in discovering meaningful latent factors using deep generative models is that the learned factors can be hard to interpret. Although causality is strongly related and can contribute to interpretability, due to the high expressiveness of the deep model, a large causal effect does not always guarantee an interpretable concept. For example, a concept that entirely replaces a letter E with a letter D, has a large causal effect on the prediction. However, such a concept does not provide valuable knowledge and is hard to interpret. To avoid such concepts, we allow the user to implement their preference or prior knowledge as an interpretability regularizer to constrain the generative model's expressive power. The proposed method then seeks for binary concepts with large causality under the constrained search space.

The integration can easily be done via a scoring function $r(\x_{\gamma_i=0}, \x_{\gamma_i=1})$ which evaluates the usefulness of concept $m_i$. Here, $\x_{\gamma_i=0}$ and $\x_{\gamma_i=1}$ are obtained from the generative model by performing the do-operation $do(\gamma_i = 0)$ and $do(\gamma_i = 1)$ on input $\x$, respectively. In this study, we introduce two regularizers which are based on the following intuitions. First, an interpretable concept should only affect a small amount of input features (Eq. \eqref{eq:compactness}). This desiderata is general and can be applied to many tasks. The second one is more task-specific in which we focus on the gray-scale image classification task. An intervention of a concept should only add or substract the pixel value, but not both at the same time (Eq. \eqref{eq:directional}). Furthermore, we desire that $\gamma_i=1$ indicates the \textit{presence} of pixels and $\gamma_i=0$ indicates the \textit{absence} of pixels. We show the detailed formulation for these regularizers in Appendix \ref{appendix:regularizer}. Using these interpretability regularizer, we observed a significant improvement in interpretability (Subsection \ref{ch:experiment-quantitative}) and consistency (Appendix \ref{appendix:regularizer}) of concepts.

\section{Experiment\label{ch:experiment}}
\subsection{Experiment setting\label{ch:experiment-setting}}
We demonstrate our method using three datasets: EMNIST\cite{Cohen2017-gv}, MNIST\cite{LeCun2010-wy} and  Fashion-MNIST\cite{Xiao2017-tk}. For each dataset, we select several classes and train a classifier on the selected classes. In particular, we select the letters `A, B, C, D, E, F' for EMNIST, digits `1, 4, 7, 9' for MNIST, and `t-shirt/top, dress, coat' for the Fashion-MNIST dataset. We note that our setting is more challenging than the common test setting in existing works (e.g., classifier for MNIST 3 and 8 digits) since a larger number of classes and concepts are involved in the classification task. Due to the space limit, here we mainly show the visual explanation obtained for the EMNIST dataset in which we use $M=3$ concepts. The dimension of $\bfalpha_i$ and $\bfbeta$ are $K=1$ and $L=7$, respectively. The explanation results of other datasets and further detailed experiment settings can be found in Appendix \ref{appendix:experiment}, \ref{appendix:mnist} and \ref{appendix:fmnist}.

\subsection{Qualitative results\label{ch:experiment-qualitative}}
\begin{figure}[t]
  \centering
  \begin{subfigure}[b]{0.6\linewidth}
    \centering
    \includegraphics[width=0.85\linewidth]{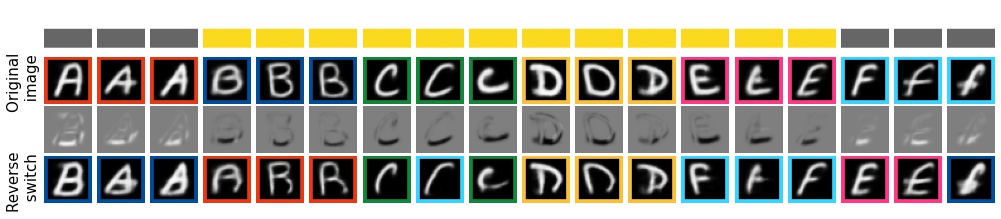}
    \caption{Controlling switch $\gamma_0$\label{fig:discrete-0} of concept $m_0$ (bottom stroke)}
  \end{subfigure}%
  \begin{subfigure}[b]{0.25\linewidth}
    \centering
    \includegraphics[width=0.5\linewidth]{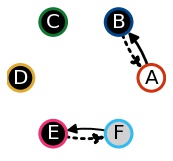}
    \caption{Transition by $m_1$\label{fig:transition-0}}
  \end{subfigure}%

  \begin{subfigure}[b]{0.6\linewidth}
    \centering
    \includegraphics[width=0.85\linewidth]{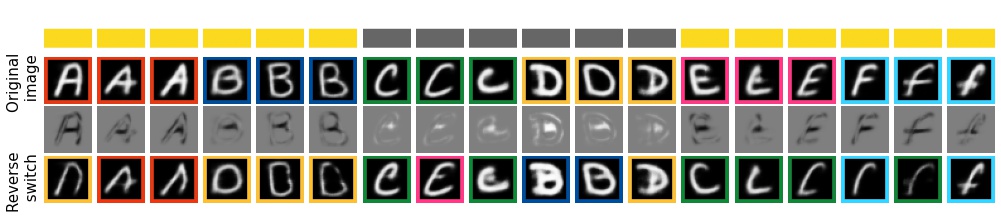}
    \caption{Controlling switch $\gamma_1$\label{fig:discrete-1} of concept $m_1$ (middle stroke)}
  \end{subfigure}%
  \begin{subfigure}[b]{0.25\linewidth}
    \centering
    \includegraphics[width=0.5\linewidth]{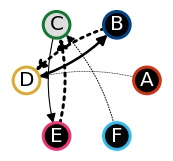}
    \caption{Transition by $m_1$\label{fig:transition-1}}
  \end{subfigure}%

  \begin{subfigure}[b]{0.6\linewidth}
    \centering
    \includegraphics[width=0.85\linewidth]{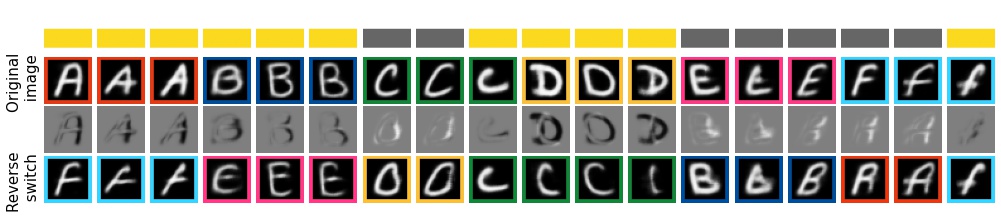}
    \caption{Controlling binary $\gamma_2$\label{fig:discrete-2} of concept $m_2$ (right stroke)}
  \end{subfigure}%
  \begin{subfigure}[b]{0.25\linewidth}
    \centering
    \includegraphics[width=0.5\linewidth]{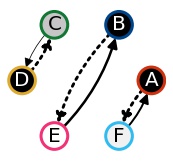}
    \caption{Transition by $m_2$\label{fig:transition-2}}
  \end{subfigure}%
  \caption{(a, c, e) The binary explanation with the intervention for each concept. (1st row) The encoded concept switch $\hat{\gamma_i}$ (yellow/gray for 1/0). (2nd row) the original reconstruction $\hat{x}$. (4th row) The reconstruction after alternating switch $\gamma_i$. (b, d, f) The transition graph of prediction output. \label{fig:binary-emnist}}
  \vspace{-1.5em}
\end{figure}

In Figure \ref{fig:binary-emnist} (\subref{fig:discrete-0}, \subref{fig:discrete-1} and \subref{fig:discrete-2}), we showed three discovered binary concepts for the EMNIST dataset. In each image, we show in the first row the encoded binary switch of concept $m_i$ for different samples, in which yellow indicates $\hat{\gamma_i}=1$ and gray indicates $\hat{\gamma_i}=0$. The second row shows the original reconstructed image $\hat{x}$ while the fourth row shows the image reconstructed when we reverse the binary switch $\hat{x}^{[i]}$. The border color indicates the prediction result of each image. Finally, the third row show the difference of $\hat{x}^{[i]}$ and $\hat{x}^{[i]}$.

We observed that the proposed method was able to discover useful binary concepts for explaining the classifier. First, the binary switches of these concepts have a large causal effect on the classifier output, i.e., alternating the switch affects the prediction. For example, Figure \ref{fig:discrete-0} explains that adding a bottom stroke to letter A has a significant effect on the classifier output. Not only that, each concept captured a group of similar interventions and can be easily interpreted, i.e., concept $m_0$ represents the bottom stroke, concept $m_1$ represents the right stroke, and concept $m_2$ represents the middle stroke.

The explanation in Figure \ref{fig:binary-emnist} (\subref{fig:discrete-0}, \subref{fig:discrete-1} and \subref{fig:discrete-2}) can be considered as a local explanation which focus on explaining specific samples. Not only that, the proposed method also excels in providing organized knowledge about the discovered concepts and prediction classes. In particular, we can aggregate the causal effect of these local explanation for each concept and class to assess how the each a binary switch change the prediction. Letting $X_u = \{\x \in X \mid f(\hat{\x}) = u\}$, the transition probability from $y=u$ to $y=v$ for a concept $m_i$ using the do operation $do(\gamma_i = d)$ ($d \in \{0, 1\}$) can be obtained as $w_{u,v}^{do(\gamma_i = d)} = \text{Pr}[y = v \mid y = u, do(\gamma_i = d)] = \frac{1}{|X_u|} \sum_{\x \in X_u} \mathbbm{1} [f(\hat{\x}^{do(\gamma_i = d)}) = v]$.

\begin{figure}[t]
  \captionsetup{aboveskip=5pt}
  \captionsetup[subfigure]{aboveskip=2pt}
  \centering
  \begin{subfigure}[b]{0.2\linewidth}
    \centering
    \includegraphics[width=0.9\linewidth]{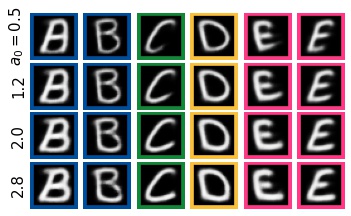}
    \caption{$\bfalpha_2$ (\textit{bottom})\label{fig:continuous-0}}
  \end{subfigure}%
  \begin{subfigure}[b]{0.2\linewidth}
    \centering
    \includegraphics[width=0.9\linewidth]{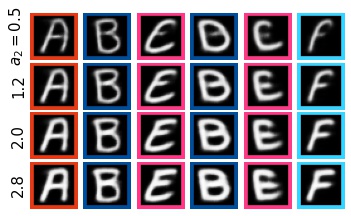}
    \caption{$\bfalpha_1$ (\textit{middle})\label{fig:continuous-1}}
  \end{subfigure}%
  \begin{subfigure}[b]{0.2\linewidth}
    \centering
    \includegraphics[width=0.9\linewidth]{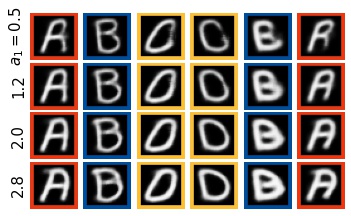}
    \caption{$\bfalpha_2$ (\textit{right})\label{fig:continuous-2}}
  \end{subfigure}%
  \begin{subfigure}[b]{0.2\linewidth}
    \centering
    \includegraphics[width=0.9\linewidth]{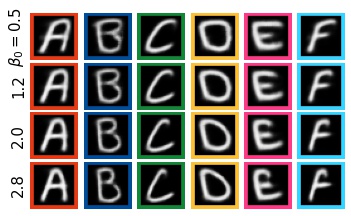}
    \caption{$\beta_6$ (\textit{width})\label{fig:continuous-3}}
  \end{subfigure}%
  \begin{subfigure}[b]{0.2\linewidth}
    \centering
    \includegraphics[width=0.9\linewidth]{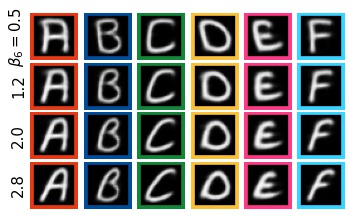}
    \caption{$\beta_6$ (\textit{skewness})\label{fig:continuous-4}}
  \end{subfigure}%
  \caption{Visualization of the learned concept-specific and global variants. The proposed method captured the variant within each causal concept, i.e., the change of shape of (a) the bottom stroke, (b) the middle stroke and (c) the right stroke. (d, e) Our method was also able to disentangle the concepts variants with other variants that does not affect the prediction.\label{fig:continuous-emnist}}
  \vspace{-1em}
\end{figure}

In \ref{fig:binary-emnist} (\subref{fig:transition-0}, \subref{fig:transition-1} and \subref{fig:transition-2}), we show the calculated transition probabilities for each concept as a graph in which each note represents a prediction class. A solid arrow (dashed arrow) represents the transition when activating (deactivating) a concept and the arrow thickness shows the transition probability $w_{u,v}^{do(\gamma_i = 1)}$ ($w_{u,v}^{do(\gamma_i = 0)}$). We neglect the transition which transition probability is less than 0.1 For example, from Figure \ref{fig:transition-0}, one can interpret that the bottom stroke is important to distinguish (E,F) and (A,B).

Finally, in Figure \ref{fig:continuous-emnist} (\subref{fig:continuous-0}, \subref{fig:continuous-1} and \subref{fig:continuous-2}), we show the captured variants within each concept and other global variants, that have a small affect on the classifier output. In contrast to binary switches, these variants explain what does not change the prediction. We first activate the concept $m_i$ using the do-operation $\text{do}(\gamma_i = 1)$, then plot the reconstruction while alternating $\bfalpha_i$. We observed that $\bfalpha_0$ captured the length of the bottom stroke, $\bfalpha_1$ captured the shape of the right stroke, and $\bfalpha_2$ captured the length of the inside (middle) stroke, respectively. Especially, our method was also able to differentiate the concept-specific variants with other global variants $\bfbeta$ such as skewness or width (Figure \ref{fig:continuous-emnist} \subref{fig:continuous-3}, \subref{fig:continuous-4}).

\subsection{Comparing with other methods.\label{ch:experiment-comparison}}
\begin{figure}[t]
  \captionsetup{aboveskip=5pt}
  \captionsetup[subfigure]{aboveskip=2pt}
  \centering
  \begin{subfigure}[b]{0.5\linewidth}
    \centering
    \includegraphics[width=0.95\linewidth]{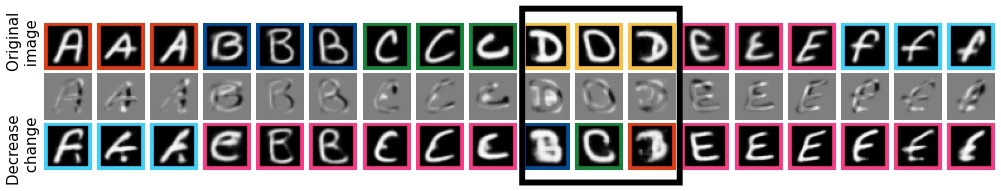}
    \caption{Intervention by decreasing $\alpha_0$}\label{fig:ccVAE-emnist-decrease}
  \end{subfigure}%
  \begin{subfigure}[b]{0.5\linewidth}
    \centering
    \includegraphics[width=0.95\linewidth]{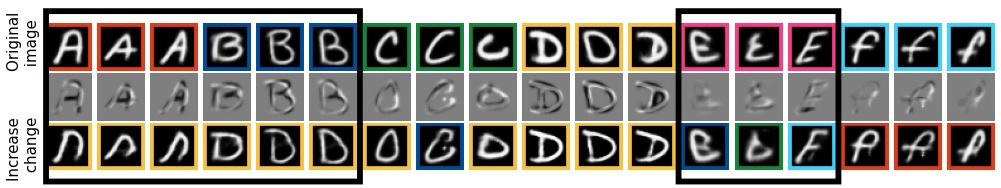}
    \caption{Intervention by increasing $\alpha_0$}\label{fig:ccVAE-emnist-increase}
  \end{subfigure}%
  \caption{A causal factor by \cite{OShaughnessy2020-yz}. Low interpretability results are framed (More details in text)\label{fig:ccVAE-emnist}}
  \vspace{-1em}
\end{figure}

We compare our method to other baselines in Figure \ref{fig:demonstration}. First, saliency-map-based methods, which use a saliency map to quantify the importance of (super)pixels, although is easy to understand, do not explain why highlighted (super)pixels are important (Figure \ref{fig:saliency-maps}). Because they only provide one explanation for each input, they can not explain how these pixels distinguish the predicted class from others classes. Our method, can provide multiple explanations by interventing difference concepts.
\begin{figure}[t]
  \centering
  \begin{subfigure}[b]{0.4\linewidth}
    \centering
    \includegraphics[width=0.7\linewidth]{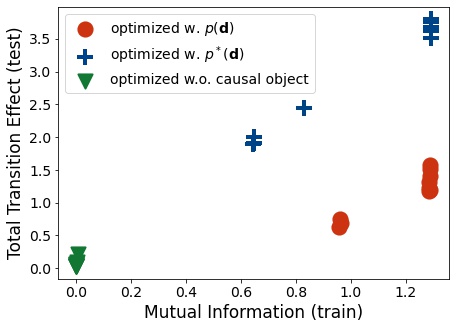}
    \caption{$p(\bfgamma)$ vs $p^*(\bfgamma)$ vs \textit{no causal}\label{fig:compare-mi}}
  \end{subfigure}%
  \begin{subfigure}[b]{0.3\linewidth}
    \centering
    \includegraphics[width=0.49\linewidth]{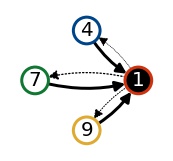}
    \includegraphics[width=0.49\linewidth]{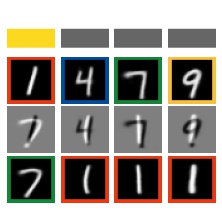}
    \caption{$\lambda_\text{R} = 0$\label{fig:mnist-no-reg}}
  \end{subfigure}%
  \begin{subfigure}[b]{0.3\linewidth}
    \centering
    \includegraphics[width=0.49\linewidth]{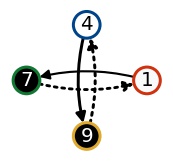}
    \includegraphics[width=0.49\linewidth]{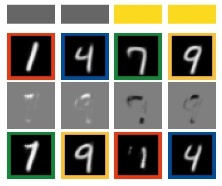}
    \caption{$\lambda_\text{R} = 1$\label{fig:mnist-with-reg}}
  \end{subfigure}%
  \caption{(a) (MNIST) Train-time MI and test-time TTE of ten runs when $\mathcal{L}_\text{CE}$ is based on $p(\bfgamma)$ (red), $p^*(\bfgamma)$ (blue), and when trained without $\mathcal{L}_\text{CE}$ (green). (b) Discovered binary concepts and their transition graph when trained with and without $\mathcal{L}_\text{R}$.}
  \vspace{-1.5em}
\end{figure}

Next, we compare to \citet{OShaughnessy2020-yz}, in which we used a VAE model with ten continuous factors and encouraged three factors to have causal effects on predicted classes. In Figure \ref{fig:ccVAE-emnist}, we visualize $\alpha_0$ which achieved the largest causal effect. In Figure \ref{fig:ccVAE-emnist-decrease} (\ref{fig:ccVAE-emnist-increase}), we decrease (increase) $\alpha_0$ until the its prediction label changes and show that intervention result in the third row. First, we observed that it failed to disentangle different causal factors as $\alpha_0$ affects all the bottom, middle and right strokes. For example, in Figure \ref{fig:ccVAE-emnist-decrease}, decreasing $\alpha_ t$ changed the letter D in the 10th column to letter B (\textit{middle stroke} concept), while changed the letter D in the 11th column to letter C (\textit{left stroke} concept). A similar result is also observed in Figure \ref{fig:ccVAE-emnist-increase} for letter E. Second, it failed to disentangle the concept-specific variant, which does not affect the prediction. For example, for the letter A and B (1st to 6th column) in Figure \ref{fig:ccVAE-emnist-increase}, increasing $\alpha_0$ does not only affect the occurrence of the \textit{middle stroke}, but also changes the shape of the \textit{right stroke}.

Our method overcomes these limitations with a carefully designed binary-discrete structure coupled with the proposed causal effect and interpretability regularizer. By encouraging the causal influence of only the binary switches, our method can disentangle what affects the prediction and the variant of samples with the same prediction. Thus, it encourages that a binary switch $m_i$ only changes the prediction from a class $y_k$ to only one other class $y_{k'}$, resulting in a more interpretable explanation. We also emphasize that the binary-continuous mixed structure alone is not enough to obtain valuable concepts for explanation (Figure \ref{fig:mixed-vae}).

\subsection{Quantitative results\label{ch:experiment-quantitative}}
We evaluate the causal influence of a concept set using the total transition effect (TTE) which is defined as $\text{TTE} = \frac{1}{M} \sum_{i \in [M]} \sum_{u,v \in [T]} [w_{u,v}^{do(\gamma_i = 1)} + w_{u,v}^{do(\gamma_i = 0)}]$ where $M$ and $T$ are the number of concepts and classes, respectively. Here, a large value of TTE indicates a significant overall causal effect by the whole discovered concept set on all class transitions. Compared to information flow, TTE can evaluate more directly and faithfully the causal effect of binary switches on dataset $X$. Moreover, it is also more easy for end-user to understand.

In Figure \ref{fig:compare-mi}, we show the test-time mutual information and the TTE values when the causal objective $\mathcal{L}_\text{CE}$ uses the prior $p^*(\bfgamma)$ (Eq. \eqref{eq:estimate-prior}), VAE model's prior $p(\bfgamma)$ and when trained without $\mathcal{L}_\text{CE}$. The interpretability regularizers are included in all settings. We observed that when $p(\bfgamma)$ is used, there are cases where the estimated mutual information is high, but the total transition effect is small. On the other hand, the mutual information obtained with estimated $p^*(\bfgamma)$ aligns better with the TTE value. We claim that this is because of the deviation between $p(\bfgamma)$ and the `true' $p^*(\bfgamma)$. By estimating $p^*(\bfgamma)$ on the run, our method can better evaluate and optimize the causal influence of $\bfgamma$ on $y$. Moreover, we also observed that without the causal objective, we failed to discover causal binary concepts.

Next, we evaluate how implementing user's preferences and prior knowledge via $\mathcal{L}_\text{R}$ increases the interpretability of concepts. In Figure \ref{fig:mnist-no-reg}, we show an example of concepts discovered when we train the model without the interpretability regularizer. We see that alternating the binary switch of this concept (top) only replaces the digit 4, 7, 9 by the digit 1 but does not provide any proper explanation why the image is identified as 1. Although this concept has a large causal effect, it barely offers valuable knowledge. Our method, using the interpretability regularizers, can discover binary concepts with high interpretability that adequately explain that digit 7 can be distinguished from digit 1 based on the existence of the top stroke (Figure \ref{fig:mnist-with-reg}).

\section{Conclusion}
We introduced the problem of discovering binary concepts for explaining a black-box classifier. We first proposed a structural generative model that can properly express binary concepts. Then, we proposed a learning process that simultaneously learns the data distribution and encourages the binary switches to have a large causal effect on the classifier output. The proposed method also allows integrating user's preferences and prior knowledge for better interpretability and consistency. We demonstrated that the proposed method could discover interpretable binary concepts with a large causal effect which can effectively explain the classification model for multiple datasets.

\begin{ack}
\end{ack}

\bibliographystyle{plainnat}
\bibliography{reference}

\appendix
\section{Appendix}
\subsection{Algorithm for estimating $I(\bfd; Y)$ \label{appendix:algorithm}}
\begin{figure}[t]
  \centering
  \begin{subfigure}[b]{0.49\linewidth}
    \centering
    \includegraphics[width=0.75\linewidth]{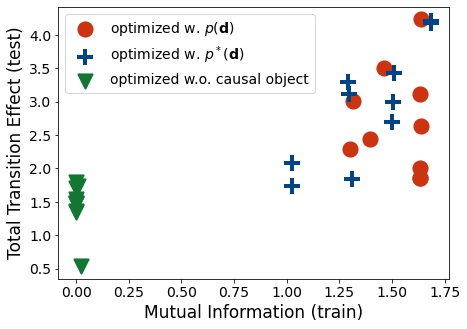}
    \caption{EMNIST}
  \end{subfigure}
  \begin{subfigure}[b]{0.49\linewidth}
    \centering
    \includegraphics[width=0.75\linewidth]{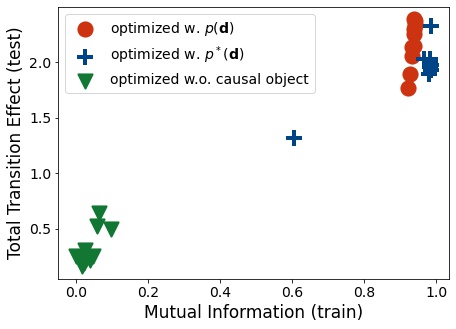}
    \caption{Fashion-MNIST}
  \end{subfigure}
  \caption{Train-time MI and test-time TTE of ten runs when $\mathcal{L}_\text{CE}$ is based on $p(\bfgamma)$ (red), $p^*(\bfgamma)$ (blue), and when trained without $\mathcal{L}_\text{CE}$ (green). Addition results for EMNIST and FMNIST. \label{fig:compare-mi-additional}}
\end{figure}%

\begin{algorithm}[h]
  \KwData{mini-batch data $X$, \\\# samples for continuous latents $N_c$, \\\# classes $T$}
  $I \gets 0$\;
  $S \gets$ \text{all combinations of $M$ binary latents}\;
  $p_y \gets zeros(T)$\;
  \For{$\s \in S$}{
    Estimate $p^*(\s)$ using Eq. \eqref{eq:prior_estimate}\;
    $p_{y \mid \s} \gets zeros(K)$\;
    \For{$j = 1$ \text{ to } $N_c$}{
      $\bm{\beta} \gets L$-dim vector from  $\mathcal{N}^*_\delta(0, I)$\;
      $\forall_i, \bm{\alpha}_i \gets M$-dim vector from $\mathcal{N}^*_\delta(0, I)$\;
      $A \gets (\bm{\alpha}_1, \bm{\alpha}_2, \dots, \bm{\alpha}_M)$\;
      $\hat{\x} \gets $ \text { sample from } $p(\x \mid \s, A, \bm{\beta})$\;
      $p_{y \mid \s} \gets p_{y \mid \s} + \frac{1}{N_c} p(y \mid \hat{\x})$\;
    }
    $I \gets I + p^*(\s) \sum_{t=1}^{T} p(y \mid \hat{\x})[t] \log p(y \mid \hat{\x})[t]$\;
    $p_y \gets p_y + p^*(\s) p(y \mid \hat{\x})$\;
  }
  $I \gets I - \sum_{t=1}^{T} p_y[t] \log p_y[t]$\;
  \Return{$I$}\;
  \caption{Algorithm for estimating $I(\bfd; Y)$.}\label{al:estimate-mi}
\end{algorithm}
Let $S = \{\s^{(0)}, \s^{(1)}, \dots, \s^{(2^M)}\}$ where $s^{(i)} \in \{0, 1\}^M$ be the set of $2^M$ possible combinations of the switches for $M$ concepts, we can obtain $\mathcal{L}_\text{CE}$ as
\begin{align}
  \mathcal{L}_\text{CE}(X) = &-\sum_{\s^{(j)}\in S} p^*(\s^{(j)}) (\sum_{y'} p(y \mid \s^{(j)})\log p(y \mid \s^{(j)})) \nonumber \\
                             &+ \sum_{y} p(y) \log p(y), \label{eq:causal_objective}
\end{align}
in which
\begin{align}
  p^*(\bfgamma) &\approx \frac{1}{|X|}\sum_{\x \in X} \prod_{i=1}^M q(\gamma_i \mid \x). \label{eq:prior_estimate}
\end{align}
Here, $q(\gamma_i \mid \x) = \pi_i \gamma_i + (1 - \pi_i) (1 - \gamma_i)$ where $(\pi_1, \dots, \pi_M) = Q^d(\x)$.

Moreover, in Eq. \eqref{eq:causal_objective}, $p(y \mid \s)$ is estimated by using $N_c$ samples of $A$ and $\bm{\beta}$ drawing from the corresponding VAE's prior $\{p(\bfalpha_i)\}_{i=1}^M$ and $p(\bm{\beta})$, respectively. The detailed algorithm is described in Algorithm \ref{al:estimate-mi}. We also show the additional result of the test-time mutual information and the TTE values when the causal object is based on estimated $p^*(\bfgamma)$ (blue), VAE model's prior $p(\bfgamma)$ (red) and when trained without causal objective (green) for EMNIST and FMNIST dataset.

\subsection{Proofs\label{appendix:proof}}
\begin{figure}[h]
  \centering
  \begin{subfigure}[b]{0.3\linewidth}
    \centering
    \includegraphics[width=1\linewidth]{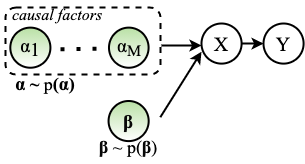}
  \end{subfigure}%
  \caption{Causal DAG of O’Shaughnessy et al. (2020)}\label{fig:existing_dag}
\end{figure}

\begin{proposition}[(O’Shaughnessy et al. (2020), Proposition 2)]\label{pro:mi-as-iflow-existing}
  The information flow from $\bfalpha$ to Y in the DAG of Figure \ref{fig:existing_dag}, in which $(\bfalpha, \bfbeta)$ are independence, coincides with the mutual information between $\bm{\alpha}$ and $Y$. That is,
  \begin{align}
    I(\bfalpha \rightarrow Y) = I(\bfalpha; Y) = \E_{\bfalpha, Y} \left[\frac{p(\bfalpha)p(Y \mid \bfalpha)}{p(\bfalpha)p(Y)}\right]
  \end{align}
\end{proposition}

\subsubsection{Proof of Proposition 1}
\begin{proof}
  Considering the proposed DAG in Figure 3a, we first let $\bfalpha' = \bfgamma$ and $\bfbeta' = (A, \bfbeta)$. Since, $\bfgamma' \sim p^*(\bfgamma)$, $(\bfalpha_1, \dots, \bfalpha_M) \sim \prod_{i=1}^Mp(\alpha_i) \text{ with } p(\alpha_i) = \mathcal{N}^\text{fold}_\delta(0; \mathcal{I})$ and $\bfbeta \sim \mathcal{N}^\text{fold}_\delta(0; \mathcal{I})$, we have that in our causal DAG, $\bfalpha'$ and $\bfbeta'$ are independence. Thus, the proposed DAG coincides with the causal DAG proposed of O’Shaughnessy et al. (2020), in which $\bfalpha$ (of Figure \ref{fig:existing_dag}) is replaced by $\bfalpha'$ and $\bfbeta$ (of Figure \ref{fig:existing_dag}) is replaced by $\bfbeta'$. From Proposition \ref{pro:mi-as-iflow-existing}, we can conclude that $I(\bfgamma \rightarrow Y)$ coincides with $I(\bfgamma; Y)$ in the proposed DAG.
\end{proof}

\subsection{On the $\delta\text{-Shifted}$ Folded Normal Distribution $\mathcal{N}^\text{fold}_\delta(\mu, \sigma^2)$\label{appendix:fold-shilf}}
\subsubsection{Why not $\mathcal{N}(\mu, \sigma^2)$ but $\mathcal{N}^\text{fold}_\delta(\mu, \sigma^2)$}
We discuss why the standard Gaussian distribution, which is a common choice for VAE models, is not appropriate for finding the proposed causal effect. Reminding that in the conceptual latent code $\z$, the corresponding code $\bm{\psi}_i$ of concept $m_i$ is $\bm{\psi}_i = \hat{\gamma_i} \times \bfalpha_i$. If $\bfalpha_i \sim \mathcal{N}(0, I)$, then for any $\bfalpha_i$ around the center zero, $\bm{\psi}_i \mid \hat{\gamma_i}=0$ and $\bm{\psi}_i \mid \hat{\gamma_i}=1$ would takes a very similar values. Thus, $p(y \mid \gamma_i = 0, \bfalpha_i)$ would be close to $p(y \mid \gamma_i = 1, \bfalpha_i)$ and it conflicts with our causal effect which requires that $p(y \mid \gamma_i = 0)$ to be different from $p(y \mid \gamma_i = 1)$ regardless value of $\bfalpha_i$. To resolve this conflict, we propose to use the $\delta\text{-Shifted}$-Folded Normal Distribution to parameterize $Q^c(A, \bm{\beta} \mid \x)$ to avoid $\bfalpha_i$ around $0$. Here, $\delta\text{-Shifted}$ Folded Normal Distribution $\mathcal{N}^\text{fold}_\delta(\mu, \sigma^2)$ is the distribution of $|x| + \delta$ for $\delta > 0$ in which $x \sim \mathcal{N}(\mu, \sigma ^2)$. We observed from our experiments that this design choice significantly boost the discovering ability for binary concepts that have large causal effect.

\subsubsection{Implementing $\mathcal{N}^\text{fold}_\delta(\mu, \sigma^2)$}
We implement the $\delta\text{-Shifted}$ Folded Normal Distribution $\mathcal{N}^\text{fold}_\delta(\mu, \sigma^2)$ using the standard implementation of the Normal Distribution $\mathcal{N}(\mu, \sigma^2)$. In particular, to obtain a sample $z$ from $\mathcal{N}^\text{fold}_\delta(\mu, \sigma^2)$, we first sample $\z' \sim \mathcal{N}(\mu, \sigma^2)$ then apply the transformation $z = |z'| + \alpha$. Since obtaining exact KL-divergence for the Folded Normal Distribution is complicated, we substitute $KL(\mathcal{N}^\text{fold}_\delta(\mu, \sigma^2), \mathcal{N}^\text{fold}_\delta(0, 1))$ by the KL-divergence of the corresponding Normal Distribution, i.e., $KL(\mathcal{N}(\mu, \sigma^2), \mathcal{N}(0, 1))$ to optimizing the VAE objective (Eq. \ref{eq:all}). Through the experiments, we observed that implementing $\mathcal{N}^\text{fold}_\delta(\mu, \sigma^2)$ this way does not cause harmful effect on the learning process. We leave the investigation of more sophisticated methods for feature work.

\subsubsection{Efficacy of using $\mathcal{N}^\text{fold}_\delta(\mu, \sigma^2)$}
\begin{figure}[h]
  \centering
  \begin{subfigure}[b]{1.\linewidth}
    \centering
    \includegraphics[width=0.45\linewidth, valign=t]{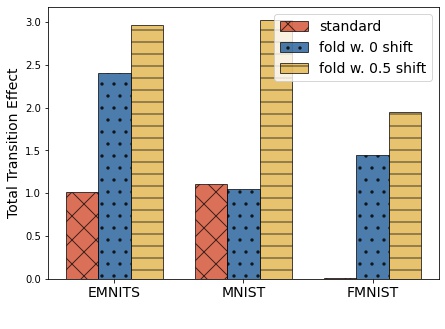}
  \end{subfigure}%
  \caption{Total Transition Effect for different choice of distribution for continuous latents.\label{fig:compare-continuous}}
  \vspace{-1.5em}
\end{figure}

We then confirm the impact of $\delta\text{-Shifted}$ Folded Normal Distribution $\mathcal{N}^\text{fold}_\delta(\mu, \sigma^2)$, and the impact of using the estimated distribution $p^*(\bfgamma)$ in evaluating the causal effect. We conduct the learning process ten times for each setting and evaluate the averaged TTE value for ten runs. In Figure \ref{fig:compare-continuous}, we show the average TTE value for different distribution choices for continuous latents. We observed that adopting the $\delta\text{-Shifted}$-Folded Normal Distribution has a significant effect in discovering causal binary concepts.

\subsection{Details of interpretability regularizers\label{appendix:regularizer}}
\begin{align}
  \mathcal{L}_\text{compact}(\x) = &\frac{1}{M}\sum_{i = 1}^M\frac{1}{P}\|\hat{\x} -\hat{\x}^{[i]}\|, \label{eq:compactness}\\
  \mathcal{L}_\text{directional}(\x) = &\frac{1}{M}\sum_{i = 1}^M\frac{1}{P} \sum_{p=1}^P \mathbbm{1}[\hat{\x}^{[i]}_p > \hat{\x}_p] \times |\hat{\x}_p - \hat{\x}^{[i]}_p| \times \hat{\gamma_i} \label{eq:directional} \\
  + &\mathbbm{1}[\hat{\x}^{[i]}_p \leq \hat{\x}_p] \times |\hat{\x}_p - \hat{\x}^{[i]}_p| \times (1 - \hat{\gamma_i}), \nonumber
\end{align}
where $M$ is the number of concepts, $P$ is the dimension of the input and $\hat{\x}^{[i]}$ is the reconstruction after \textit{reversing} the latent code $\hat{\gamma}_i$ of concept $m_i$. We give a brief interpretation for Eq. \eqref{eq:directional}. Consider a concept $m_i$ in a sample $\x$. If concept $m_i$ is activated, i.e., $\hat{\gamma}_i=1$, then $\hat{\x}^{[i]}$ corresponds to the \textit{turn off} intervention $do(\gamma_i = 0)$. In this case, we expect that this intervention only removes some pixels in $\hat{x}$. Thus, we penalize the difference $|\hat{\x}_p - \hat{\x}^{[i]}_p|$ for positions $p$ where the pixel value increases, i.e., where $\hat{\x}^{[i]}_p > \hat{\x}_p$. Finally, we combine these regularizers as $\mathcal{L}_\text{R}(\x) = \lambda_3 \mathcal{L}_\text{compact}(\x) + \lambda_4 \mathcal{L}_\text{directional}(\x)$.

\subsubsection{Inconsistency issue of discovered concept}
\begin{figure}[h]
  \centering
  \begin{subfigure}[b]{1.0\linewidth}
    \centering
    \includegraphics[width=0.45\linewidth]{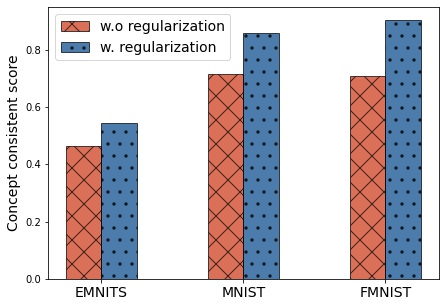}
  \end{subfigure}%
  \caption{The consistence score of discovered concepts across successful runs when being trained with and without the interpretability regularizer.\label{fig:regularization-effect}}
\end{figure}
Training the generative model with interpretability regularizers also resolves to an extent the inconsistency issue of discovered concepts, i.e., each trial results in a different set of concepts. Since only humans can confirm if the concepts are meaningful, this inconsistency can lead to a cherry-picking scheme that is, users might keep training the model until he/she encounts favorite concepts accidentally. We note that this is also a common problem of explanation methods that use deep generative models. We evaluate the consistency between different runs based on the similarity of the prediction transition graphs. Letting $I, I'$ be two arbitrary permutations of $\{1, 2, \dots, M\}$, we evaluate the similarity of two transition graphs $W$ and $W'$ as follows
\begin{align}
  \text{Similarity}(W, W') = & \max_{I, I'} \text{CosSim}(W_{(I)}, W'_{(I')}) \\
  \text{where} \quad
                             & W_{(I)} = (W_{i_0}, W_{i_1}, \dots, W_{i_M})_{i_j \in I} \nonumber \\
                             & W'_{(I')} = (W'_{i'_0}, W'_{i'_1}, \dots, W'_{i'_M})_{i'_j \in I'} \nonumber \\
  \text{and} \quad &W_{i} = (w_{u,v}^{do(\gamma_i =0)}, w_{u,v}^{do(\gamma_i = 1)})_{u, v \in [T]} \nonumber
\end{align}
Here, the maximal operation $\max_{I, I'}$ is used to consider different orders of the discovered concepts, which can be arbitrary in each run. In Figure \ref{fig:regularization-effect}, we show the average similarities between different runs for each dataset. To avoid the cases in which the optimization fails, we only consider those runs which achieved a large causal effect (i.e., best four runs in ten runs). We observed that adding the regularization also improves the consistency of the discovering concepts. In this work, we demonstrated our method with grayscale images, where small parts in the image take an essential role in classifying the object. We leave the exploration of other regularizations for other domains to future work.

\subsection{Experiment details\label{appendix:experiment}}
\begin{table}[h]
\centering
\caption{Classifier architecture}\label{tab:classifier-network}
\begin{tabular}{r|r|r}
  Layer (type)       & Output Shape     & \# Param \\ \hline
            Conv2d-1 & [-1, 32, 16, 16] & 544      \\
            Conv2d-3 & [-1, 64, 8, 8]   & 32,832   \\
            Linear-6 & [-1, 64]         & 262,208  \\
           Linear-10 & [-1, 256]        & 16,640   \\
           Linear-11 & [-1, 256]        & 16,640   \\
           Linear-14 & [-1, 4]          & 1,028    \\
  Linear-15          & [-1, 4]          & 1,028    \\ \hline
  Total params       &                  & 330,920
\end{tabular}
\end{table}

\begin{table}[h]
\centering
\caption{Generative model architecture}\label{tab:vae-network}
\begin{tabular}{r|r|r}
  Layer (type)       & Output Shape     & \# Param \\ \hline
  \multicolumn{3}{c}{Encoder}                    \\\hline
            Conv2d-1 & [-1, 64, 16, 16] & 1,024    \\
            Conv2d-3 & [-1, 128, 8, 8]  & 131,072  \\
            Conv2d-5 & [-1, 256, 4, 4]  & 524,288  \\
            Conv2d-7 & [-1, 64, 4, 4]   & 16,384   \\
            Conv2d-9 & [-1, 2, 1, 1]    & 2,050    \\
           Conv2d-10 & [-1, 10, 1, 1]   & 10,250   \\
           Conv2d-11 & [-1, 10, 1, 1]   & 10,250   \\
           Conv2d-12 & [-1, 8, 1, 1]    & 8,200    \\
  Conv2d-13          & [-1, 8, 1, 1]    & 8,200    \\ \hline
  \multicolumn{3}{c}{Decoder}                    \\ \hline
  ConvTranspose2d-17 & [-1, 64, 4, 4]   & 10,240   \\
  ConvTranspose2d-19 & [-1, 256, 4, 4]  & 16,384   \\
  ConvTranspose2d-21 & [-1, 128, 8, 8]  & 524,288  \\
  ConvTranspose2d-23 & [-1, 64, 16, 16] & 131,072  \\
  ConvTranspose2d-25 & [-1, 1, 32, 32]  & 1,024    \\ \hline
  Total params       &                  & 1,394,726
\end{tabular}
\end{table}

All of our experiments were run using two GeForce RTX 3090 GPUs. We use the standard split for all dataset (MNIST, EMNIST, FMNIST) which are distributed via the torchvision package. We then select the sample with targeted classes for each experiment to obtain the final train and test set. All input images are resized to $32 \times 32$ images. We show the network architecture of the classifier and the VAE model used in each experiment in Table \ref{tab:classifier-network}, \ref{tab:vae-network}. The classifier is trained with a batch size of $256$ using the Adam optimizer with a learning rate of $0.0005$ for $20$ epochs. The trained classifier achieved an accuracy of $95.45\%$ on the test dataset. The VAE model is optimized with a batch size of $265$ using the Adam optimizer with learning rate $0.0005$ for $30$ epochs. At each training step, the causal effect term is estimated using the Algorithm \ref{al:estimate-mi} with $N_c$ samples for continuous latents.

For the EMNIST experiment (class A, B, C, D, E, F), we adopted a generative model that contains $M=3$ concepts in which each concept-specific variant $\bfa_i$ has a dimension of $K=1$. The dimension of the non-causal factor $\bm{\beta}$ is set to $L=7$. We resize the input image to $32 \times 32$ grayscale image. The other hyper-parameters are set as $\lambda_1= \lambda_2=1$, $\lambda_3 = 50$, $\lambda_4=1000$ and $\lambda_{CE}=50$. Moreover, we employ the $0.5\text{-Shifted}$ Folded Normal Distribution $\mathcal{N}^\text{fold}_\delta(\mu, \sigma^2)$, i.e., $\delta=0.5$ for the continuous distribution. We train the generative model using Adam with a learning rate $lr=0.0005$ for 30 epochs.

For the MNIST experiment (class 1,4,7,9), we adopted a generative model that contains $M=2$ concepts in which each concept-specific variant $\bfa_i$ has a dimension of $K=1$. The dimension of the non-causal factor $\bm{\beta}$ is set to $L=8$. The other hyper-parameters are set as $\lambda_1= \lambda_2=1$, $\lambda_3 = 100$, $\lambda_4=1000$ and $\lambda_{CE}=50$. Specially, for the compactness regularizer coefficient $\lambda_3$, we init $\lambda_3=5$ at the start of the training process, and increase $\lambda_3$ by $5$ for each $100$ update steps until $\lambda_3=100$. Moreover, we initialize the Relaxed Bernoulli distribution's temperature at $0.4$ and decrease it using annealing schedule with annealing rate $0.0001$ with until it reaches $0.1$. In the testing phase, this temperature is set to $0$ to obtain a (non-relaxed) Bernoulli distribution.

On the other hand, for the Fashion-MNIST experiment (t-shirt/top, dress, coat), we adopted a generative model that contains $M=2$ concepts in which each concept-specific variant $\bfa_i$ has a dimension of $K=1$. The dimension of the non-causal factor $\bm{\beta}$ is set to $L=8$. The other hyper-parameters are set as $\lambda_1= \lambda_2=1$, $\lambda_3 = 10e$, $\lambda_4=1000$ and $\lambda_{CE}=25$. Other parameters are identical with the setting for MNIST dataset.

\subsection{Results for MNIST\label{appendix:mnist}}
\begin{figure*}[t]
  \centering
  \begin{subfigure}[b]{0.5\linewidth}
    \centering
    \includegraphics[width=0.3\linewidth]{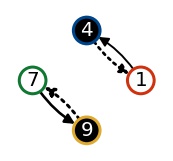}
  \end{subfigure}%
  \begin{subfigure}[b]{0.5\linewidth}
    \centering
    \includegraphics[width=0.3\linewidth]{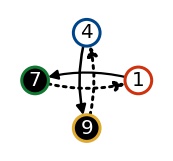}
  \end{subfigure}%

  \begin{subfigure}[b]{0.5\linewidth}
    \centering
    \includegraphics[width=0.95\linewidth]{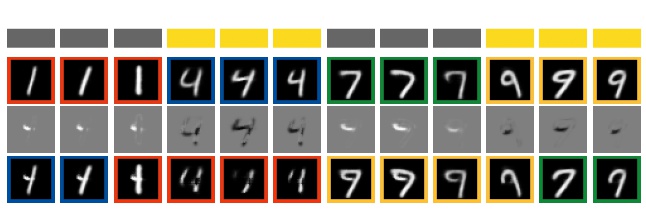}
    \caption{Controlling switch $\gamma_0$\label{fig:mnist-discrete-0} (middle stroke)}
  \end{subfigure}%
  \begin{subfigure}[b]{0.5\linewidth}
    \centering
    \includegraphics[width=0.95\linewidth]{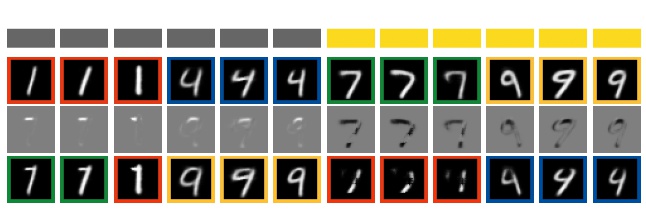}
    \caption{Controlling binary $\gamma_1$\label{fig:mnist-discrete-1} (top stroke)}
  \end{subfigure}%
  \caption{(MNIST) The binary explanation. (1st row) The encoded concept switch $\hat{\gamma_i}$ (yellow/gray for 1/0). (2nd row) the original reconstruction $\hat{x}$. (4th row) The reconstruction after alternating switch $\gamma_i$.\label{fig:binary-mnist}}
\end{figure*}

\begin{figure*}[t]
  \centering
  \begin{subfigure}[b]{0.2\linewidth}
    \centering
    \includegraphics[width=0.8\linewidth]{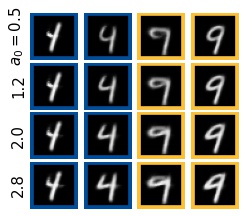}
    \caption{ $\bfalpha_0$\label{fig:mnist-continuous-0} (middle)}
  \end{subfigure}%
  \begin{subfigure}[b]{0.2\linewidth}
    \centering
    \includegraphics[width=0.8\linewidth]{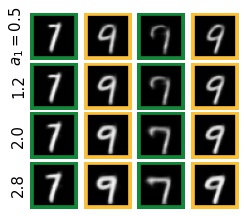}
    \caption{ $\bfalpha_1$\label{fig:mnist-continuous-1} (top)}
  \end{subfigure}%
  \begin{subfigure}[b]{0.2\linewidth}
    \centering
    \includegraphics[width=0.8\linewidth]{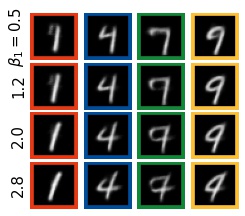}
    \caption{ $\beta_1$\label{fig:mnist-continuous-4} (cross)}
  \end{subfigure}%
  \begin{subfigure}[b]{0.2\linewidth}
    \centering
    \includegraphics[width=0.8\linewidth]{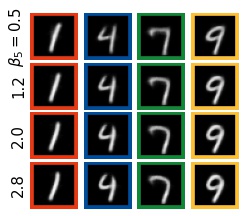}
    \caption{ $\beta_5$\label{fig:mnist-continuous-2} (roundness)}
  \end{subfigure}%
  \begin{subfigure}[b]{0.2\linewidth}
    \centering
    \includegraphics[width=0.8\linewidth]{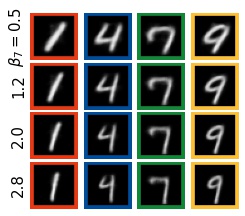}
    \caption{ $\beta_7$\label{fig:mnist-continuous-3} (width)}
  \end{subfigure}%
  \caption{(MNIST) Visualization of the learned concept-specific and global variants.\label{fig:continuous-mnist}}
\end{figure*}

\begin{figure*}[t]
  \centering
  \begin{subfigure}[b]{0.5\linewidth}
    \centering
    \includegraphics[width=0.3\linewidth]{fig/exp_562_causal_transition_0.jpeg}
  \end{subfigure}%
  \begin{subfigure}[b]{0.5\linewidth}
    \centering
    \includegraphics[width=0.3\linewidth]{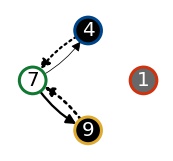}
  \end{subfigure}%

  \begin{subfigure}[b]{0.5\linewidth}
    \centering
    \includegraphics[width=0.95\linewidth]{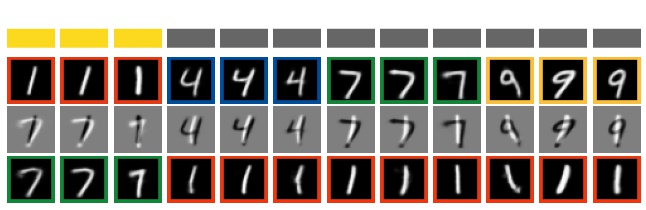}
    \caption{Controlling switch $\gamma_0$\label{fig:mnist-discrete-0-noreg}}
  \end{subfigure}%
  \begin{subfigure}[b]{0.5\linewidth}
    \centering
    \includegraphics[width=0.95\linewidth]{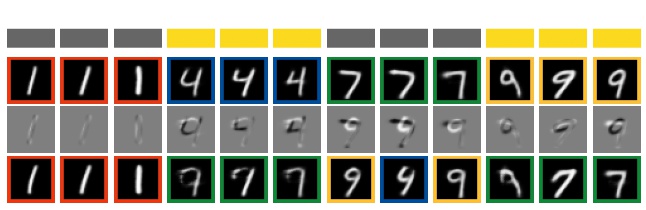}
    \caption{Controlling binary $\gamma_1$\label{fig:mnist-discrete-1-noreg}}
  \end{subfigure}%
  \caption{(MNIST without $\mathcal{L}_\text{R}$) The binary explanation.\label{fig:binary-mnist-noreg}}
\end{figure*}

\begin{figure*}[t]
  \centering
  \begin{subfigure}[b]{0.2\linewidth}
    \centering
    \includegraphics[width=0.8\linewidth]{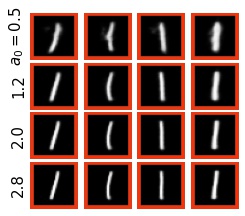}
    \caption{ $\bfalpha_0$\label{fig:mnist-continuous-0-noreg}}
  \end{subfigure}%
  \begin{subfigure}[b]{0.2\linewidth}
    \centering
    \includegraphics[width=0.8\linewidth]{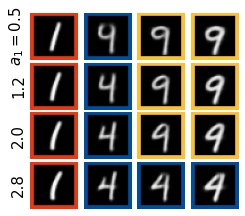}
    \caption{ $\bfalpha_1$\label{fig:mnist-continuous-1-noreg}}
  \end{subfigure}%
  \begin{subfigure}[b]{0.2\linewidth}
    \centering
    \includegraphics[width=0.8\linewidth]{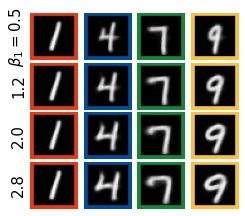}
    \caption{ $\beta_1$\label{fig:mnist-continuous-2-noreg}}
  \end{subfigure}%
  \begin{subfigure}[b]{0.2\linewidth}
    \centering
    \includegraphics[width=0.8\linewidth]{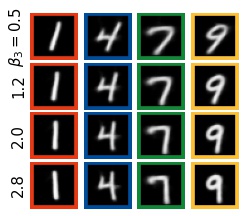}
    \caption{ $\beta_3$\label{fig:mnist-continuous-3-noreg}}
  \end{subfigure}%
  \begin{subfigure}[b]{0.2\linewidth}
    \centering
    \includegraphics[width=0.8\linewidth]{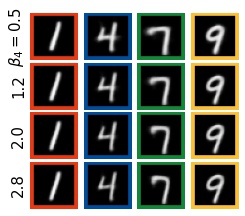}
    \caption{ $\beta_4$\label{fig:mnist-continuous-4-noreg}}
  \end{subfigure}%
  \caption{(MNIST without $\mathcal{L}_\text{R}$) Visualization of the learned concept-specific and global variants.\label{fig:continuous-mnist-noreg}}
\end{figure*}

We show the result for dataset MNIST in Figure \ref{fig:binary-mnist} and \ref{fig:continuous-mnist}. We also show the result obtained when trained without the interpretability regularizers in Figure \ref{fig:binary-mnist-noreg} and \ref{fig:continuous-mnist-noreg}.

\subsection{Results of Fashion-MNIST\label{appendix:fmnist}}
\begin{figure*}[t]
  \centering
  \begin{subfigure}[b]{0.5\linewidth}
    \centering
    \includegraphics[width=0.3\linewidth]{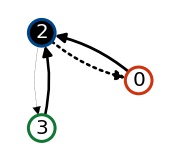}
  \end{subfigure}%
  \begin{subfigure}[b]{0.5\linewidth}
    \centering
    \includegraphics[width=0.3\linewidth]{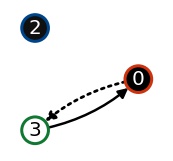}
  \end{subfigure}%

  \begin{subfigure}[b]{0.5\linewidth}
    \centering
    \includegraphics[width=0.9\linewidth]{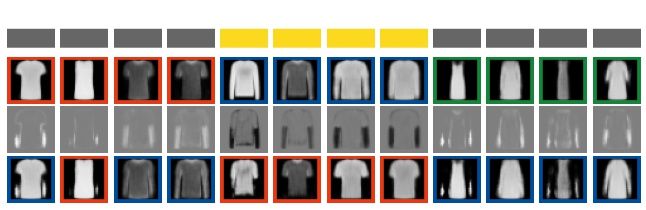}
    \caption{Controlling switch $\gamma_0$\label{fig:fmnist-discrete-0} (long sleeve)}
  \end{subfigure}%
  \begin{subfigure}[b]{0.5\linewidth}
    \centering
    \includegraphics[width=0.9\linewidth]{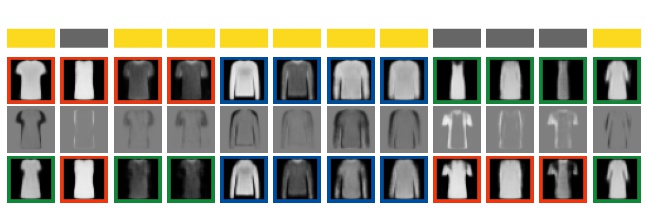}
    \caption{Controlling binary $\gamma_1$\label{fig:fmnist-discrete-1} (shoulder + body shape)}
  \end{subfigure}%
  \caption{(Fashion-MNIST) The binary explanation. (1st row) The encoded concept switch $\hat{\gamma_i}$ (yellow/gray for 1/0). (2nd row) the original reconstruction $\hat{x}$. (4th row) The reconstruction after alternating switch $\gamma_i$.\label{fig:binary-fmnist}}
\end{figure*}

\begin{figure*}[t]
  \centering
  \begin{subfigure}[b]{0.2\linewidth}
    \centering
    \includegraphics[width=0.95\linewidth]{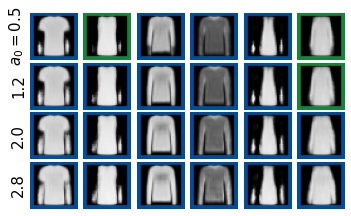}
    \caption{ $\bfalpha_0$\label{fig:fmnist-continuous-0} (long sleeve)}
  \end{subfigure}%
  \begin{subfigure}[b]{0.2\linewidth}
    \centering
    \includegraphics[width=0.95\linewidth]{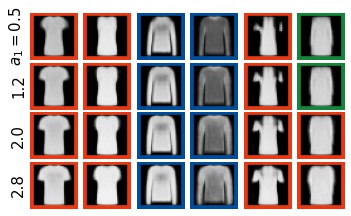}
    \caption{ $\bfalpha_1$\label{fig:fmnist-continuous-1} (shoulder)}
  \end{subfigure}%
  \begin{subfigure}[b]{0.2\linewidth}
    \centering
    \includegraphics[width=0.95\linewidth]{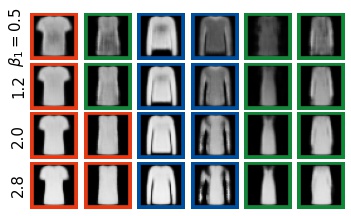}
    \caption{ $\beta_6$\label{fig:fmnist-continuous-5} (lower width)}
  \end{subfigure}%
  \begin{subfigure}[b]{0.2\linewidth}
    \centering
    \includegraphics[width=0.95\linewidth]{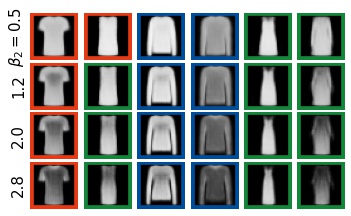}
    \caption{ $\beta_6$\label{fig:fmnist-continuous-3} (darkness)}
  \end{subfigure}%
  \begin{subfigure}[b]{0.2\linewidth}
    \centering
    \includegraphics[width=0.95\linewidth]{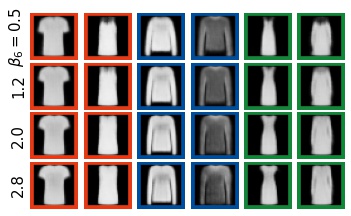}
    \caption{ $\beta_6$\label{fig:fmnist-continuous-4} (neck height)}
  \end{subfigure}%
  \caption{(Fashion-MNIST) Visualization of the learned concept-specific and global variants.\label{fig:continuous-fmnist}}
\end{figure*}

\begin{figure*}[t]
  \centering
  \begin{subfigure}[b]{0.5\linewidth}
    \centering
    \includegraphics[width=0.3\linewidth]{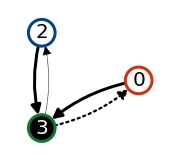}
  \end{subfigure}%
  \begin{subfigure}[b]{0.5\linewidth}
    \centering
    \includegraphics[width=0.3\linewidth]{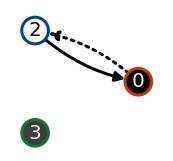}
  \end{subfigure}%

  \begin{subfigure}[b]{0.5\linewidth}
    \centering
    \includegraphics[width=0.9\linewidth]{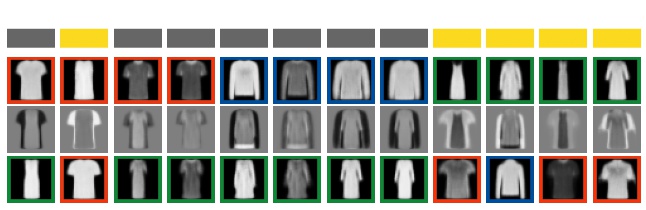}
    \caption{Controlling switch $\gamma_0$\label{fig:fmnist-discrete-0-noreg}}
  \end{subfigure}%
  \begin{subfigure}[b]{0.5\linewidth}
    \centering
    \includegraphics[width=0.9\linewidth]{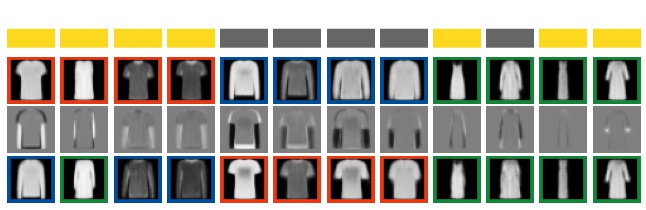}
    \caption{Controlling binary $\gamma_1$\label{fig:fmnist-discrete-1-noreg}}
  \end{subfigure}%
  \caption{(Fashion-MNIST without $\mathcal{L}_\text{R}$) The binary explanation.\label{fig:binary-fmnist-noreg}}
\end{figure*}

\begin{figure*}[t]
  \centering
  \begin{subfigure}[b]{0.2\linewidth}
    \centering
    \includegraphics[width=0.95\linewidth]{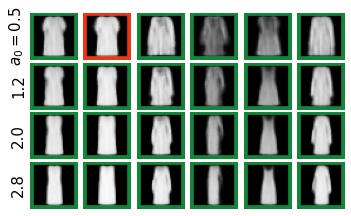}
    \caption{ $\bfalpha_0$\label{fig:fmnist-continuous-0-noreg}}
  \end{subfigure}%
  \begin{subfigure}[b]{0.2\linewidth}
    \centering
    \includegraphics[width=0.95\linewidth]{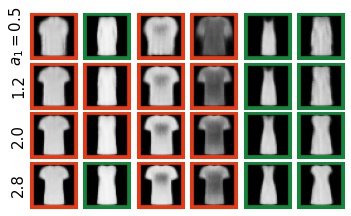}
    \caption{ $\bfalpha_1$\label{fig:fmnist-continuous-1-noreg}}
  \end{subfigure}%
  \begin{subfigure}[b]{0.2\linewidth}
    \centering
    \includegraphics[width=0.95\linewidth]{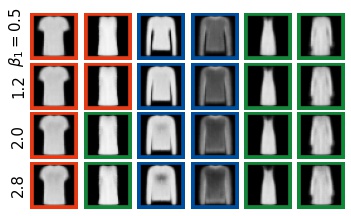}
    \caption{ $\beta_6$\label{fig:fmnist-continuous-5-noreg}}
  \end{subfigure}%
  \begin{subfigure}[b]{0.2\linewidth}
    \centering
    \includegraphics[width=0.95\linewidth]{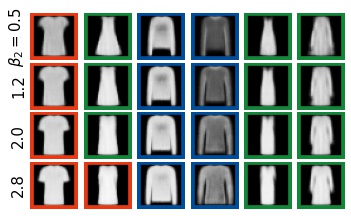}
    \caption{ $\beta_6$\label{fig:fmnist-continuous-3-noreg}}
  \end{subfigure}%
  \begin{subfigure}[b]{0.2\linewidth}
    \centering
    \includegraphics[width=0.95\linewidth]{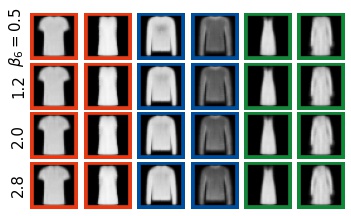}
    \caption{ $\beta_6$\label{fig:fmnist-continuous-4-noreg}}
  \end{subfigure}%
  \caption{(Fashion-MNIST without $\mathcal{L}_\text{R}$) Visualization of the learned concept-specific and global variants.\label{fig:continuous-fmnist-noreg}}
\end{figure*}

We show the result for dataset Fashion-MNIST in Figure \ref{fig:binary-fmnist} and \ref{fig:continuous-fmnist}. We also show the result obtained when trained without the interpretability regularizers in Figure \ref{fig:binary-fmnist-noreg} and \ref{fig:continuous-fmnist-noreg}.

\subsection{Addition results of EMNIST\label{appendix:emnist}}
\begin{figure*}[t]
  \centering
  \begin{subfigure}[b]{0.33\linewidth}
    \centering
    \includegraphics[width=0.6\linewidth]{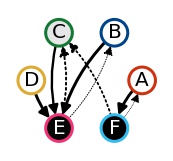}
  \end{subfigure}%
  \begin{subfigure}[b]{0.33\linewidth}
    \centering
    \includegraphics[width=0.6\linewidth]{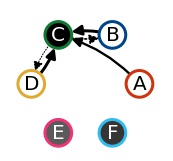}
  \end{subfigure}%
  \begin{subfigure}[b]{0.33\linewidth}
    \centering
    \includegraphics[width=0.6\linewidth]{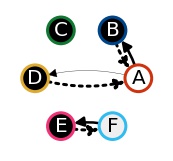}
  \end{subfigure}%

  \begin{subfigure}[b]{0.33\linewidth}
    \centering
    \includegraphics[width=0.95\linewidth]{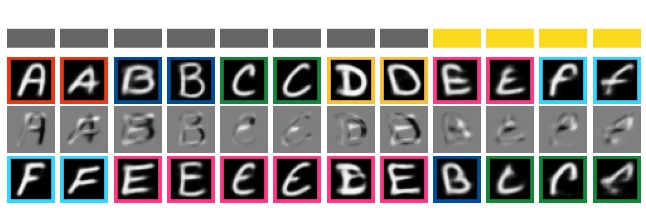}
    \caption{Controlling switch $\gamma_0$\label{fig:emnist-discrete-0-noreg}}
  \end{subfigure}%
  \begin{subfigure}[b]{0.33\linewidth}
    \centering
    \includegraphics[width=0.95\linewidth]{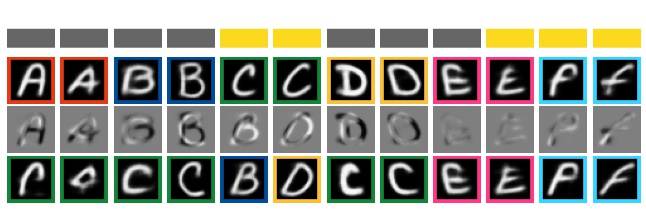}
    \caption{Controlling binary $\gamma_1$\label{fig:emnist-discrete-1-noreg}}
  \end{subfigure}%
  \begin{subfigure}[b]{0.33\linewidth}
    \centering
    \includegraphics[width=0.95\linewidth]{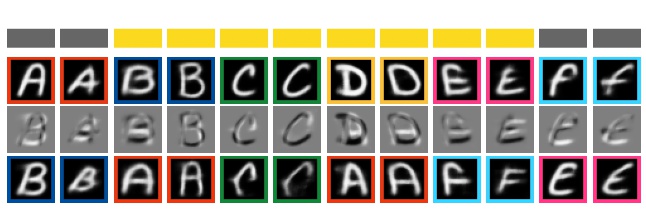}
    \caption{Controlling binary $\gamma_1$\label{fig:emnist-discrete-2-noreg}}
  \end{subfigure}%
  \caption{(EMNIST without $\mathcal{L}_\text{R}$) The binary explanation.\label{fig:binary-emnist-noreg}}
\end{figure*}

\begin{figure*}[t]
  \centering
  \begin{subfigure}[b]{0.2\linewidth}
    \centering
    \includegraphics[width=0.8\linewidth]{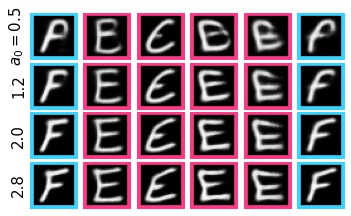}
    \caption{ $\bfalpha_0$\label{fig:emnist-continuous-0-noreg}}
  \end{subfigure}%
  \begin{subfigure}[b]{0.2\linewidth}
    \centering
    \includegraphics[width=0.8\linewidth]{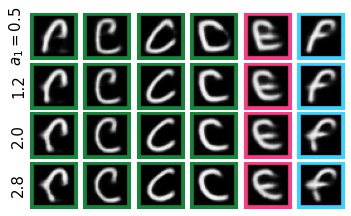}
    \caption{ $\bfalpha_1$\label{fig:emnist-continuous-1-noreg}}
  \end{subfigure}%
  \begin{subfigure}[b]{0.2\linewidth}
    \centering
    \includegraphics[width=0.8\linewidth]{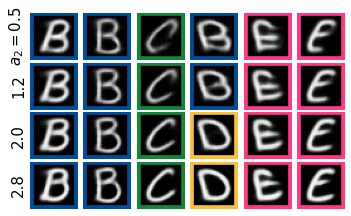}
    \caption{ $\bfalpha_2$\label{fig:emnist-continuous-2-noreg}}
  \end{subfigure}%
  \begin{subfigure}[b]{0.2\linewidth}
    \centering
    \includegraphics[width=0.8\linewidth]{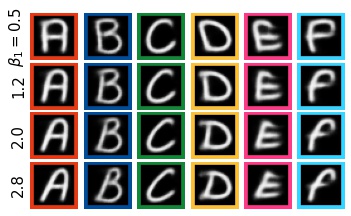}
    \caption{ $\beta_1$\label{fig:emnist-continuous-3-noreg}}
  \end{subfigure}%
  \begin{subfigure}[b]{0.2\linewidth}
    \centering
    \includegraphics[width=0.8\linewidth]{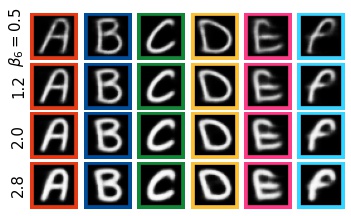}
    \caption{ $\beta_6$\label{fig:emnist-continuous-4-noreg}}
  \end{subfigure}%
  \caption{(EMNIST without $\mathcal{L}_\text{R}$) Visualization of the learned concept-specific and global variants.\label{fig:continuous-emnist-noreg}}
\end{figure*}

We show the result obtained when trained without the interpretability regularizers for EMNIST in Figure \ref{fig:binary-emnist-noreg} and \ref{fig:continuous-emnist-noreg}.

\end{document}